\documentclass[10pt,twocolumn,letterpaper]{article}

\usepackage[pagenumbers]{cvpr} %

\usepackage[accsupp]{axessibility}  %

\usepackage[pagebackref,breaklinks,hidelinks]{hyperref}

\usepackage{amsmath}
\usepackage{amssymb}
\usepackage{mathtools}
\usepackage{amsthm}

\usepackage{url}            %
\usepackage{booktabs}       %
\usepackage{amsfonts}       %
\usepackage{nicefrac}       %
\usepackage{microtype}      %
\usepackage[table,xcdraw,dvipsnames]{xcolor}
\usepackage{graphicx,epstopdf}
\usepackage{adjustbox}
\usepackage{multirow}
\usepackage{csquotes}
\usepackage{wrapfig}
\usepackage{pifont}  %
\usepackage{makecell}  %
\usepackage[numbers,sort,compress]{natbib}
\usepackage{diagbox}  %
\usepackage{gensymb}  %

\usepackage{enumitem}
\usepackage{subcaption}
\usepackage{comment}

\usepackage{mathtools}

\usepackage{xspace}
\makeatletter
\DeclareRobustCommand\onedot{\futurelet\@let@token\@onedot}
\def\@onedot{\ifx\@let@token.\else.\null\fi\xspace}

\renewcommand{\emph}[1]{\textit{{#1}}}
\def\eg{\emph{e.g}\onedot} 
\def\ie{\emph{i.e}\onedot} 
 
\def\etc{\emph{etc}\onedot} \def\vs{\emph{vs}\onedot}

\makeatother

\usepackage{sistyle}
\SIthousandsep{,}

\usepackage{upgreek}

\def\mD{\mathcal{D}}

\def\mI{\mathcal{I}}

\def\mO{\mathcal{O}}

\def\mY{\mathcal{Y}}

\def\1n{\mathbf{1}_n}
\def\0{\mathbf{0}}
\def\1{\mathbf{1}}

\def\L{{\bf L}}

\def\R{{\mathbb R}}

\def\c{{\bf c}}

\def\n{{\bf n}}

\def\p{{\bf p}}

\def\r{{\bf r}}

\def\w{{\bf w}}

\DeclareMathOperator*{\bbE}{\mathbb{E}}

\def\bbS{{\mathbb{S}}}

\newcommand{\cP}{\mathcal{P}}
\newcommand{\cS}{\mathcal{S}}

\usepackage{physics}

\usepackage{array}
\newcolumntype{C}[1]{>{\centering\arraybackslash}m{#1}}  %

\definecolor{orange}{rgb}{0.8, 0.6, 0.2}
\definecolor{lightgrey}{rgb}{0.7, 0.7, 0.7}

\newcommand{\TOPIC}[1]{\textcolor{lightgrey}{\bf [TOPIC: #1]}}
\renewcommand{\TOPIC}[1]{} %

\definecolor{rtwocolor}{RGB}{50,50,200}
\definecolor{ronecolor}{RGB}{200,50,50}
\definecolor{rthreecolor}{RGB}{50,200,50}

\usepackage[dvipsnames]{xcolor}

\newcommand{\cmark}{\color{Green}{\ding{51}}}%
\newcommand{\cmarkstar}{\color{Green}{\ding{51}$^*$}}%
\newcommand{\cmarkflat}{\color{Green}{\ding{51}$^\flat$}}%
\newcommand{\hmark}{\color{YellowOrange}{\ding{51}}}%
\newcommand{\xmark}{\color{Red}{\ding{55}}}%

\newcommand{\figsym}{Fig\onedot}
\newcommand{\equsym}{eq\onedot}

\newcommand{\figref}[1]{\figsym~\ref{#1}}
\newcommand{\equref}[1]{\equsym~\eqref{#1}}

\usepackage{placeins}
\usepackage{xr}

\newcommand{\graymidrule}{\arrayrulecolor[HTML]{DCDCDC} \midrule \arrayrulecolor[HTML]{000000}}

\renewcommand{\paragraph}[1]{\vspace{0.5em}\noindent\textbf{#1}}

\usepackage{ccicons}  %

\usepackage{float}

\usepackage[capitalize]{cleveref}
\crefname{section}{Sec.}{Secs.}
\Crefname{section}{Section}{Sections}
\Crefname{table}{Table}{Tables}
\crefname{table}{Tab.}{Tabs.}

\def\cvprPaperID{1185} %
\def\confName{CVPR}
\def\confYear{2023}

\begin{document}

\title{Pointersect: Neural Rendering with Cloud-Ray Intersection}

\author{
	Jen-Hao Rick Chang$^{1}$, 
	Wei-Yu Chen$^{1,2}$\thanks{Work done at Apple. Corresponding author: \url{jenhao_chang@apple.com}}, 
	Anurag Ranjan$^{1}$, 
	Kwang Moo Yi$^{1,3*}$, 
	Oncel Tuzel$^{1}$ \\
$^{1}$Apple, 
$^{2}$Carnegie Mellon University, 
$^{3}$University of British Columbia\\
{\small \url{https://machinelearning.apple.com/research/pointersect}}
}
\maketitle

\begin{abstract}

We propose a novel method that renders point clouds as if they are surfaces.
The proposed method is differentiable and requires no scene-specific optimization.
This unique capability enables, out-of-the-box, surface normal estimation, rendering room-scale point clouds, inverse rendering, and ray tracing with global illumination.
Unlike existing work that focuses on converting point clouds to other representations---\eg, surfaces or implicit functions---our key idea is to directly infer the intersection of a light ray with the underlying surface represented by the given point cloud. 
Specifically, we train a set transformer that, given a small number of local neighbor points along a light ray, provides the intersection point, the surface normal, and the material blending weights, which are used to render the outcome of this light ray.
Localizing the problem into small neighborhoods enables us to train a model with only 48 meshes and apply it to unseen point clouds.
Our model achieves higher estimation accuracy than state-of-the-art surface reconstruction and point-cloud rendering methods on three test sets.
When applied to room-scale point clouds, without any scene-specific optimization, the model achieves competitive quality with the state-of-the-art novel-view rendering methods. 
Moreover, we demonstrate ability to render and manipulate Lidar-scanned point clouds such as lighting control and object insertion.

\end{abstract}

\begin{figure}
	\centering
	\begin{subfigure}[t]{\linewidth}
		\centering
		\includegraphics[width=\linewidth]{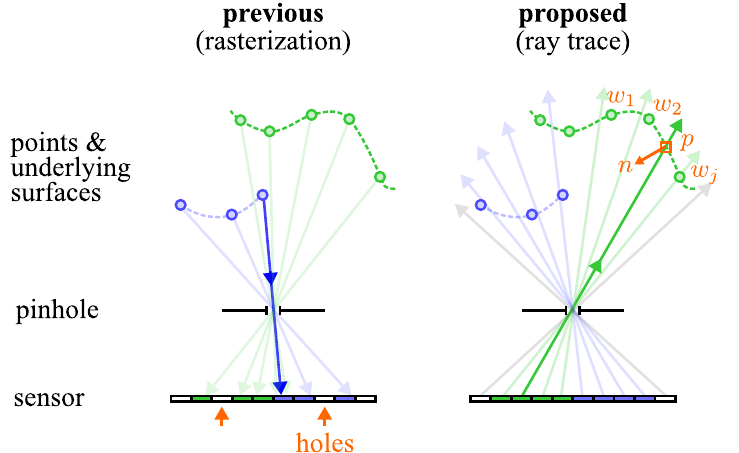}
		\caption{Illustration of the proposed method}
	\end{subfigure} 
	\\[0.85mm]
	\begin{subfigure}[t]{\linewidth}
		\centering
		\includegraphics[width=\linewidth]{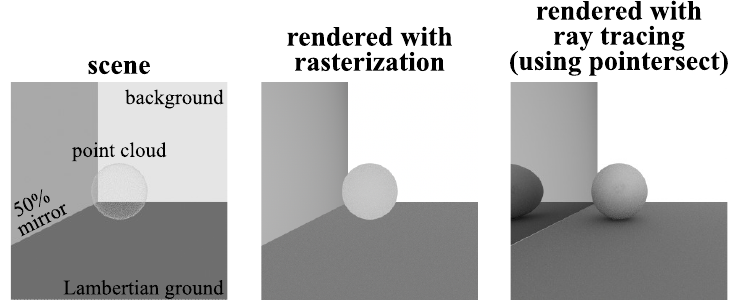}
		\caption{Rendered results of a sphere represented by points}
	\end{subfigure} 
    \vspace{-2.6mm}
	\caption{We propose \textit{pointersect}, a novel method to perform cloud-ray intersection. 
		(a) Instead of projecting points onto the sensor, suffering from holes, we trace rays from the sensor and estimate the intersection point $p$ between a ray and the underlying surface represented by the points.  We additionally estimate the surface normal $\vec{n}$, and the convex combination weights of points near the ray to blend material or color, $w_j$. 
		(b) The capability to perform cloud-ray intersection enables us to render point clouds with the standard ray tracing method, \ie, path tracing. The result shows the effect of global illumination, \eg, the cast shadow and reflection.
	}
	\label{fig:teaser}
\end{figure}

\section{Introduction}
\label{sec: intro}

\TOPIC{Point clouds are important and useful}
Point clouds are abundant.
They are samples of surfaces, and can be captured by sensors such as Lidar, continuous-wave time-of-flight, and stereo camera setups.
Point-cloud representation provides a straightforward connection to the location of the surfaces in space, and thus is an intuitive primitive to represent geometry~\cite{gross2011point}.  %

\TOPIC{Point clouds are hard to render / existing techniques}
Despite being ubiquitous, a core limitation of point clouds is that they are non-trivial to render.
Each point in the point cloud occupies no volume---one cannot render them into images as is.
Therefore, existing methods either assign each point a volume-occupying shape,  %
\eg, an oriented disk \cite{pfister2000surfels,yifan2019differentiable}, a sphere \cite{lassner2021pulsar}, or turn it into other shape representations like meshes~\cite{4408892} or implicit functions~\cite{ kazhdan2006poisson, peng2021shape,ma2021neural,hanocka2020point2mesh,feng2022np}.
However, it is difficult to determine the ideal shape, \eg, the radius of spheres or disks, for rendering. 
Small shapes would cause holes, while large shapes would cause blobby renderings, producing artifacts in the rasterized images.
While the artifacts can be alleviated by finetuning the rasterized images with an additional neural rendering step, the operation often requires per-scene training~\cite{aliev2020neural, dai2020neural, lassner2021pulsar}.
On the other hand, transforming point clouds into other shape representations complicates the pipeline and prevents gradients passing back to the point cloud through the new shape representation (in the case of inverse rendering).
For example, turning point clouds into a mesh or a Signed Distance Function (SDF)~\cite{hanocka2020point2mesh, ma2021neural} %
would require any changes on the point cloud to trigger retraining of these representations, which would clearly be prohibitive.  %

Recent works \cite{xu2022point, ost2022neural} raise new ideas to directly perform ray-casting on point clouds. 
For each scene, these methods first learn a feature embedding for each point; then they aggregate features near each camera ray to predict colors.
However, these methods require per-scene training since the feature embedding is scene-dependent. 
In our case, we aim for a solution that does not require scene-specific optimization and can be applied to any scene.

\TOPIC{What we do at a high level}

In this work, we propose \textit{pointersect}, an alternative that can directly \textit{ray-trace} point clouds by allowing one to use point clouds as surface primitives, as shown in \autoref{fig:teaser}. 
That is, we propose to train a neural network that provides the surface intersection point, surface normal, and material blending weights---the necessary information to render (or ray trace) a surface---given a point cloud and a query ray.

\TOPIC{Details}

Implementing this idea requires paying attention to details.
A core observation is that the problem to find the intersecting surface point is SE(3) equivariant---any rigid transform on the input (\ie, the point cloud and the query ray) should result in the rigid transformation of the output (\ie, the intersection point and the surface normal).
Naively training a neural network would require the network to learn this equivariance, which is non-trivial~\cite{sun2021canonical,feng2022prif,aumentado2022representing}.
Instead, we opt to remove the need for learning this equivariance by canonicalizing the input according to the queried light ray.
In addition, pointersect should be invariant to the order in which the points are provided---we thus utilize a transformer to learn the set function.

\TOPIC{Remarks on why our method shines}

It is important to note that finding intersection points between rays and surfaces is a highly atomic and localized problem which can be solved only with local information.
Thus, we design our method to only consider nearby points, where the surface would have been, and how the surface texture and normal can be derived from these nearby points.
By constraining the input to be a small number ($\sim$100) of neighboring points associated to a query ray, our method can be trained on only a handful of meshes, then be applied to unseen point clouds. 
As our experiments show, while only trained on 48 meshes, pointersect significantly improves the Poisson surface reconstruction, a scene-specific optimization method, on three test datasets.
We also demonstrate the generality and differentiability of pointersect on various applications:
novel-view synthesis on room-scale point clouds,
inverse rendering,
and ray tracing with global illumination.
Finally, we render room-scale Lidar-scanned point clouds and showcase the capability to directly render edited scenes, without any scene-specific optimization.

\TOPIC{Summary of contributions}
In short, our contributions are:
\vspace{-.5em}
\begin{itemize}[leftmargin=*]
\setlength\itemsep{-.1em}
    \item We propose \textit{pointersect}, a neural network performing the \textit{cloud-ray intersection} operation.  Pointersect is easy to train, and once learned, can be applied to unseen point clouds---we evaluate the same model on three test datasets.
    \item We demonstrate various applications with pointersect, including room-scale point cloud rendering, relighting, inverse rendering, and ray tracing.
    \item We apply pointersect on Lidar-scanned point clouds and demonstrate novel-view synthesis and scene editing.
\end{itemize}
\vspace{-.5em}

We encourage the readers to examine results and videos of novel-view rendering, relighting, and scene editing in the supplemental material and website (\url{https://machinelearning.apple.com/research/pointersect}). %

\section{Related work}
This section briefly summarizes strategies to render point clouds. 
Please see \autoref{table: related works} in the supplementary for a detailed overview on the capabilities/limitations of each method.

\paragraph{Rasterizing point clouds.}
Rasterization is a common method to render point clouds.
The idea is to project each point onto the sensor while making sure closer points occlude farther points.
Filling holes between points is the key problem in point-cloud rasterization. 
Classical methods like visibility splatting~\cite{pfister2000surfels, yifan2019differentiable} cover holes by replacing points with oriented disks. 
However, since the size and shape of the holes between projected points depend on the distribution of the points in space, the method cannot fill in all gaps. 
Recently, \citet{zhang2022differentiable} achieve high-quality results via alpha-blending surface splatting~\cite{zwicker2001surface}, coarse-to-fine optimization, and point insertion. However, their method requires per-scene optimization on the point cloud.

Recent works propose to combine rasterization with neural networks.
Given a point cloud and RGB images of a scene, \citet{aliev2020neural} and \citet{ruckert2022adop} learn an embedding for each point by rasterizing feature maps and minimizing the difference between the rendered images and the given ones.
The hole is handled by downsampling then upsampling the rasterized feature map.
Similarly, \citet{dai2020neural} aggregate points into multi-plane images, combined with a 3D-convolutional network.
\citet{huang2022boosting} use a U-Net refinement network to fill in the holes.
Recently, \citet{rakhimov2022npbg++} show that point embedding can be directly extracted from input images, which enables their method to render unseen point clouds without per-scene optimization.

It is difficult to render global illumination with these methods, since rasterization does not support such an operation.
Moreover, we show empirically in \Cref{sec: exp} that our method renders higher quality images than recent methods while directly working on $xyz$ and $rgb$, without a feature extractor.

\paragraph{Converting into other representations.}
An alternative way to render a point cloud is converting it to other primitives, such as an indicator function~\cite{kazhdan2006poisson,peng2021shape}, 
a SDF~\cite{ma2021neural}, or a mesh~\cite{hanocka2020point2mesh}.
Poisson surface reconstruction~\cite{kazhdan2006poisson, kazhdan2013screened} fits an indicator function (\ie, 1 inside the object and 0 otherwise) to the input point cloud by solving a Poisson optimization problem. 
Meshes can then be extracted from the learned function, allowing ray tracing to be performed.
However, Possion surface reconstruction requires vertex normal, which can be difficult to estimate even when the points only contain a small amount of noise, and it is difficult to support non-watertight objects.
\citet{peng2021shape} learn the indicator function with a differentiable solver and incorporate Poisson optimization in a neural network.

\citet{ma2021neural} learn a SDF represented as a neural field, and \citet{hanocka2020point2mesh} fit a deformable mesh to an objects with a self-prior. 
These methods require per-scene optimization, and to the best of our knowledge, have not been extended to render surface colors.

Alternatively, \citet{feng2022np} propose a new primitive, Neural Point, or a collection of local neural surfaces extracted from the point cloud. 
The method supports new scenes and estimates surface normal; however, dense sampling or cube marching is needed to render novel views, and the method does not render color.

\paragraph{Ray-casting.}
Instead of rasterization, point clouds can also be rendered by ray-casting. 
Early work \cite{adamson2003approximating, adams2005efficient, wald2005interactive, guennebaud2007algebraic, kolluri2008provably, amenta2004defining} develop iterative algorithms or formulate optimization programs to intersect a ray with the approximated local plane constructed by nearby points.
These methods rely on neighborhood kernels, which is non-trivial to determine~\cite{gross2011point,kobbelt2004survey}.
Recently, \citet{xu2022point} and \citet{ost2022neural} learn feature embedding at each point and aggregate the features along a query ray. 
\citet{xu2022point} march camera rays through the point cloud, average neighbor features, predict density and color by a Multi-Layer Perceptron (MLP), and render the final color via volumetric rendering~\cite{mildenhall2021nerf}. 
\citet{ost2022neural} utilize a transformer to aggregate point features into a feature associated with the query ray and predict color by an MLP. 
Both methods do not estimate surface intersection points or normal, and they need per-scene training~\cite{ost2022neural} or per-scene fine-tuning~\cite{xu2022point} to achieve high-quality results.
As we show later, our method can be applied to completely novel classes of scenes without any retraining.

\paragraph{Neural rendering from 2D images.}
Recently, Neural Radiance Field (NeRF) and similar methods \cite{mildenhall2021nerf, zhang2021nerfactor, yariv2020multiview, riegler2021stable,sitzmann2021light, wang2021ibrnet, lassner2021pulsar, esposito2022kiloneus, diolatzis2022active, lyu2022neural} have demonstrated high-quality novel-view synthesis results.
Due to the lack of immediately available 3D information, most of these methods require a per-scene optimization or surface reconstruction.
In this work, we focus on rendering a point cloud, where 3D information is available, without additional per-scene training.

\section{Method}
\label{sec: pointersect}

We aim to directly perform ray casting with point clouds. 
We thus first introduce the generic surface-ray intersection. %
We then introduce how we enable cloud-ray intersection and discuss how it can be used for actual rendering.

\subsection{Problem formulation}

\paragraph{Surface-ray intersection.}
Surface-ray intersection is a building-block operator in physics-based rendering \cite{pharr2016physically}. 
At a high level, it identifies the contacting point between a query ray and the scene geometry so that key information like material properties, surface normal, and incoming light at the point can be retrieved and computed \cite{pharr2016physically}.
Most graphics primitives allow the intersecting point to be easily found.  
For example, with triangular meshes we simply intersect the query ray with individual triangles (while using accelerated structures to reduce the number of triangles of interest). 
However, for point clouds, finding the intersection point becomes a challenging problem.

\paragraph{Cloud-ray intersection.} 
We formulate the cloud-ray intersection as follows.  
We are given the following information:  
\vspace{-1.8em}
\begin{itemize}[leftmargin=*]
	\setlength\itemsep{-.1em}
	\item a set of points $\cP = \{p_1, \dots, p_n\}$ that are samples on a surface $\cS$, where $p_i \in \R^3$ is $i$-th point's coordinate;
    \item optionally, the material (\eg, RGB color or Cook-Torrance coefficients \cite{cook1982reflectance}) associated with each point, $c_i \in \R^d$;
	\item a querying ray $\r = (r_o, \vec{r_d})$, where $r_o \in \R^3$ is the ray origin and  $\vec{r_d} \in \bbS^2$ is the ray direction.
\end{itemize}
Our goal is to estimate the following quantities: 
\vspace{-.5em}
\begin{itemize}[leftmargin=*]
	\setlength\itemsep{-.1em}
    \item the probability, $h \in [0, 1]$, that $\r$ intersects with the underlying surfaces of $\cP$; 
	\item the intersection point $p \in \R^3$ between $\r$ and $\cS$;
	\item the surface normal $\vec{n}$ at $p$;
	\item and optionally the blending weights $\w = [w_1, \dots, w_n]$, 
    where $w_i \in [0, 1]$ 
    and $\sum_{i=1}^{n} w_i \, {=} \, 1$, to estimate the material property at $p$: $c(p) = \sum_{i=1}^{n} w_i \, c_i$.
\end{itemize}
Note that this problem is under-determined---there can be infinitely many surfaces passing through points in $\cP$. 
Thereby, hand-crafting constraints such as surface smoothness are typically utilized to solve the problem \cite{kobbelt2004survey}.
In this work, we learn surface priors from a dataset of common objects using a neural network, so manual design is not needed.

\subsection{Pointersect}

Pointersect is a neural network, $f_\theta(\r, \cP) \mapsto (h, p, \vec{n}, \w)$, that estimates the intersection point and surface normal between $\r$ and the underlying surface of $\cP$.
We learn the network by formulating a regression problem, using a dataset of meshes. 
We generate training data by randomly sampling rays and point clouds on the meshes, \ie, our inputs, and running a mesh-ray intersection algorithm~\cite{pharr2016physically} to acquire the ground-truth outputs.

Despite the simplicity of the framework, we note, however, %
care must be taken when designing $f_\theta$.
Specifically, we incorporate the following geometric properties into the design of $f_\theta$ to ease training and allow generalization:

First, we utilize the fact that the intersection point is along the query ray and design $f$ to estimate the distance $t$ from the ray origin, \ie, $p = r_o + t \, \vec{r_d}$.  %

Second, we utilize the SE(3) equivariance---rotating and translating the ray and the points together should move the intersection point in the same way.
Mathematically, for all rotation matrix $R \in$ SO(3) and translation $b \in \R^3$, we have 
\begin{equation*}
	f_\theta(\r, \cP) \mapsto (h, t, \vec{n}, \w)  {\Rightarrow} f_\theta(T(\r), T(\cP)) \mapsto  (h, t, R\vec{n}, \w), 
\end{equation*}
where $T(\cP) {=} \{R\, p_i {+} b\}_{i=1\dots n}$ and $T(\r) {=} (T(r_o), R\,\vec{r_d})$.
Additionally, this holds for any point along the ray, thus
\begin{equation*}
	f_\theta(\r, \cP) {\mapsto} (h, t, \vec{n}, \w)  {\Rightarrow} f_\theta((r_o + t' \vec{r_d}, \vec{r_d}) , \cP) {\mapsto}  (h, t-t', \vec{n}, \w), 
\end{equation*}
for all $0 \le t' \le t$.
Thereby, to eliminate this ambiguity,
given a query ray $\r = (r_o, \vec{r_d})$, we rotate and translate both the ray and the scene such that $R \, \vec{r_d} = (0, 0, 1)$ and the closest point in the half space defined by $\r$ has $z=0$.

Third, the intersection point can be estimated by using the local neighborhood of the ray.
Given $\r$ and $\cP$, we form a new set of points, $\cP_\r$, by keeping only the closest $k$ points (in terms of their orthogonal distances to $\r$) within a cylinder of radius $\delta$ surrounding the ray.

\begin{figure}[t]
	\centering
	\includegraphics[width=\linewidth]{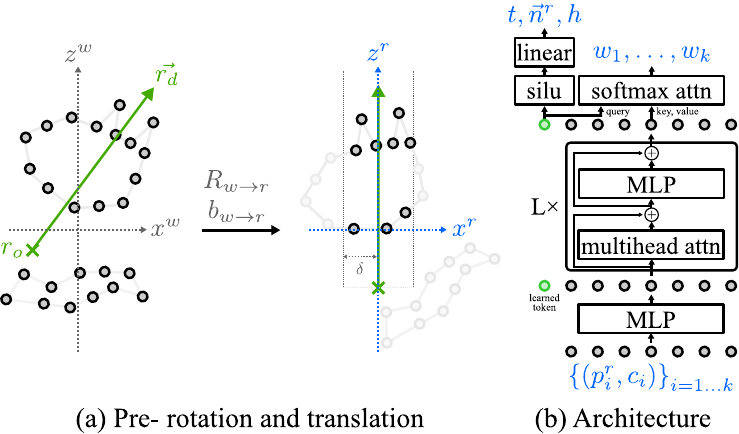}
	\vspace{-6mm}
	\caption{The pointersect model $f$ is composed of a pre-processing step and a transformer. (a) The pre-processing step rotates and translates the world coordinate such that the ray lies on the $z^r$-axis. The new coordinate origin is selected such that the closest point to the ray origin lies on the $z^r = 0$ plane. (b) The transformer takes as inputs the points in the new coordinate, \ie, $(p_i^r, c_i)$.  The output of a learned token is used to estimate $t$, the traveling distance between the ray origin and the intersection point, $\vec{n}^r$, the surface normal in the new coordinate, and $h$, the output of a sigmoid function on a logit (not drawn) to predict the probability that the ray hits a surface. It is also used as the query in the softmax attention to calculate the weight $\w$. Finally, we transform $\vec{n}^r$ back to the world coordinate and make sure it point to the opposite direction of the ray. The transformer has 4 layers ($L=4$) and a feature dimension of 64.
    }
	\label{fig: network}
\end{figure}

Last, since $\cP$ is a set, \ie, the order of the points in $\cP$ is irrelevant, we use a set transformer \cite{vaswani2017attention} as our architecture of choice.  
See \autoref{fig: network} for an overview and \cref{sec: architecture details} for detailed descriptions.

We train $f_\theta$ by optimizing the following problem: 
\begin{align}
	\min_\theta \, \bbE_{\r, \cP} \, \hat{h} \left( \lambda \| t - \hat{t} \|_2^2 +  \| \vec{n} \cross \hat{\vec{n}} \|^2_2 + \| c - \hat{c} \|_1 \right)  \nonumber \\ 
    + \hat{h} \log h + (1 - \hat{h}) \log (1 - h), \label{eq: training loss}
\end{align}
where $\hat{t}$, $\hat{\vec{n}}$, $\hat{c}$, and $\hat{h} \in \{0, 1\}$ are the ground-truth ray traveling distance, surface normal, color, and ray hit, respectively, and  $c = \sum  w_i \, c_i$ is the output color.
The expectation is over the query rays and the point clouds, which we sample randomly every iteration from a dataset of meshes.
We omit the dependency on $\cP$ and $\r$ in the notations for simplicity.

\subsection{Rendering point clouds with pointersect}
\label{sec: render}

Our ability to perform cloud-ray intersection allows two main techniques to render a point cloud.

\paragraph{Image-based rendering.}
Given a point cloud $\cP = \{ (p_i, c_i)\}_{i=1 \dots n}$, where each point has both position $p_i \in \R^3$ and RGB color $c_i \in \R^3$, and the target camera intrinsic and extrinsic matrices, we can simply cast camera rays toward $\cP$. 
The final color of a camera ray, $c(\r)$ can be computed by
\begin{equation}
	c(\r) = \sum_{i=1}^{n}  w_i(\r, \cP) \, c_i,
	\label{eq: blending}
\end{equation}
where $w_i(\r, \cP)$ is the blending weight of $p_i$ estimated by $f_\theta$, and $w_i(\r, \cP) = 0$ if $p_i$ is not a neighbor point of $\r$.

\paragraph{Rendering with ray tracing.}
A unique capability of pointersect is ray tracing.
Ray tracing allows occlusion and global illumination effects like cast shadow and specular reflection to be faithfully rendered.
Suppose we are given a point cloud $\cP = \{ (p_i, c_i)\}_{i=1 \dots n}$, where each point has both position $p_i \in \R^3$ and material information $c_i \in \R^d$ like albedo, Cook-Torrance \cite{cook1982reflectance}, or emission coefficients, and we have the environment map and the target camera intrinsic and extrinsic matrices.
We can use standard ray tracing techniques like path tracing \cite{james1986rendering} to render the image. 
At a high level, we trace a ray through multiple intersections with the point cloud until reaching the environment map or background.
At each intersection point, we calculate the material property by interpolating the material of neighboring points (using the blending equation \eqref{eq: blending}), shade the intersection point based on the reflection equation \cite{james1986rendering}, and determine the direction to continue tracing \cite{shirley2018ray}.

When we end the ray tracing with a single bounce, \ie, using pointersect to determine the first intersection point with the scene, and directly shade the intersection point with the environment, we get an equivalent algorithm of rasterization with deferred shading for meshes \cite{deering1988triangle}.

\section{Experiments}
\label{sec: exp}

We evaluate the proposed method's capability to estimate intersection points and surface normal. 
We then showcase the use of pointersect in various applications, including rendering point clouds with image-based rendering and ray-tracing, and inverse rendering.
Note that the same model is used for all experiments---no per-scene optimization is used.

\subsection{Model training}

\begin{figure}[t]
	\vspace{-1mm}
	\centering
	\begin{subfigure}{0.6\linewidth}
		\centering 
		\includegraphics[width=\linewidth]{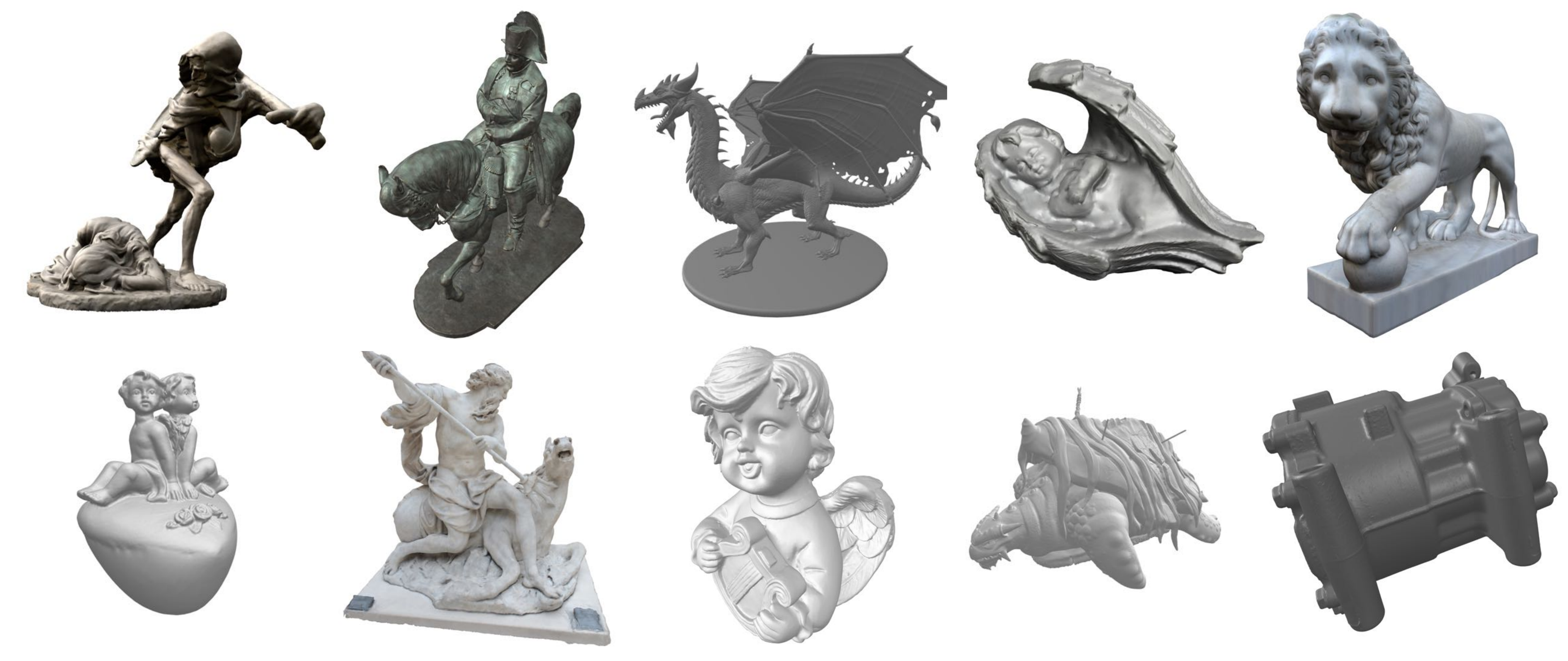}
	\end{subfigure}
\hspace{2mm}
	\begin{subfigure}{0.25\linewidth}
		\centering 
		\includegraphics[width=\linewidth]{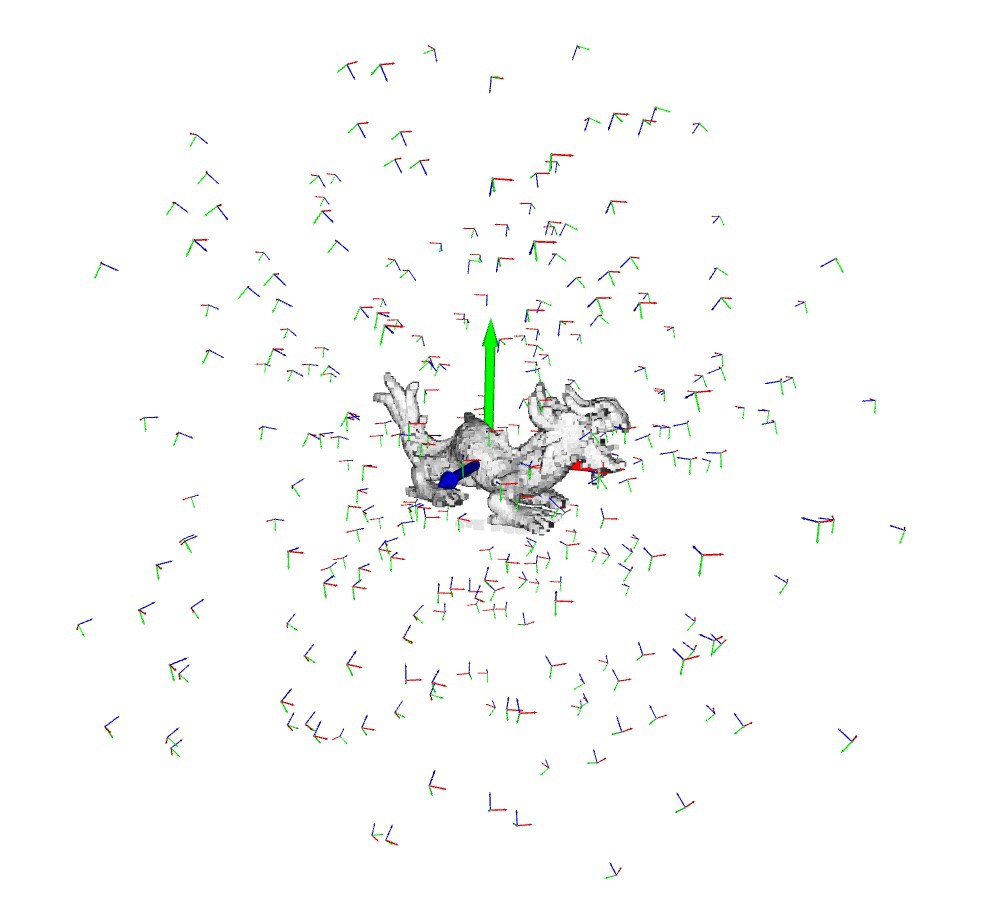}
	\end{subfigure}
	\vspace{-3mm}
	\caption{ Training meshes (left) and example camera poses (right).}
	\label{fig: train mesh and pose}
	\vspace{-2mm}
\end{figure}

The model is trained on 48 training meshes in the sketchfab dataset \cite{qian2020pugeo}.%
We show 10 training meshes in \autoref{fig: train mesh and pose} and all 48 training meshes and their download links and credits in \autoref{fig: sketchfab composition} in \autoref{sec: sketchfab dataset}.
The meshes are centered and scaled such that the longest side of their bounding box is 2 units. 
For each training iteration, we randomly select one mesh and construct 30 input cameras and 1 target camera, which capture RGBD images using the mesh-ray intersection method in Open3D \cite{Zhou2018} without anti-aliasing filtering.
We design the input and target camera poses to be those likely to appear in novel-view synthesis. 
Specifically, we uniformly sample camera position within a spherical shell of radius 0.5 to 3, looking at a random position in the unit-length cube containing the object, as shown in \autoref{fig: train mesh and pose}.
During testing, the camera poses are non-overlapping with the training poses ---  the poses used in \autoref{table: basic}, shown in \autoref{fig: basic}, have a spiral trajectory with radius changing periodically between 3 to 4 and those used in room scenes (\autoref{fig: hypersim} and \ref{fig: arkitscenes}) are chosen to follow the room layouts.

The input RGBD images create the input point cloud, where each point carries only point-wise information, including $xyz$, $rgb$, and the direction from input camera to the particular point.
To support point clouds without these information, we randomly drop $rgb$ and other features independently 50 \% of the time---during inference, we use only $xyz$ and $rgb$ for all experiments.
We also use a random $k \in [12, 200]$ at every iteration.
To help learning blending weights, at every iteration we select a random image patch from ImageNet dataset as the texture map for the mesh.
We train the model for 350,000 iterations, and it takes 10 days on 8 A100 GPUs.
Please see more details in \autoref{sec: model details}.

\subsection{Evaluating pointersect}
\label{sec: exp basics}

\begin{figure*}[t]
	\centering
	\begin{subfigure}[t]{\linewidth}
		\centering
		\includegraphics[width=\linewidth]{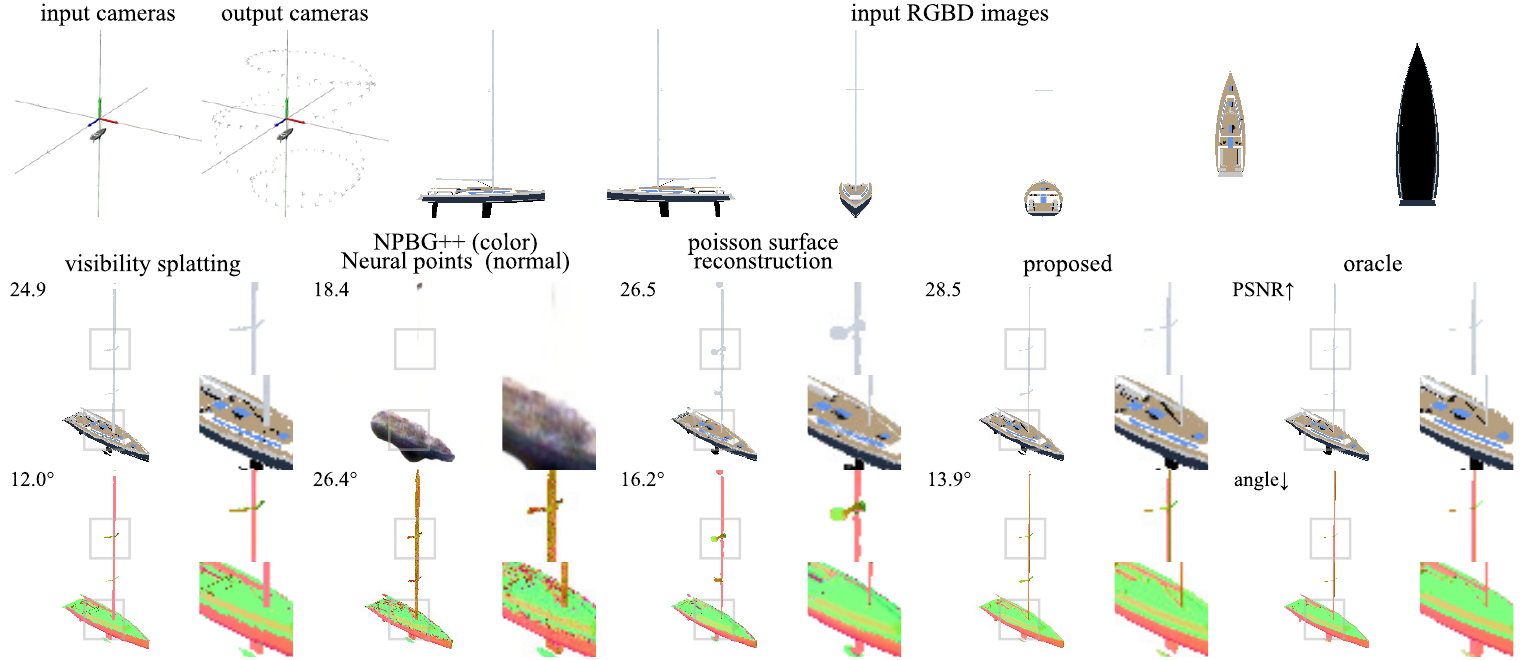}
	\end{subfigure}
	\\[1em]
	\begin{subfigure}[t]{\linewidth}
		\centering
		\includegraphics[width=\linewidth]{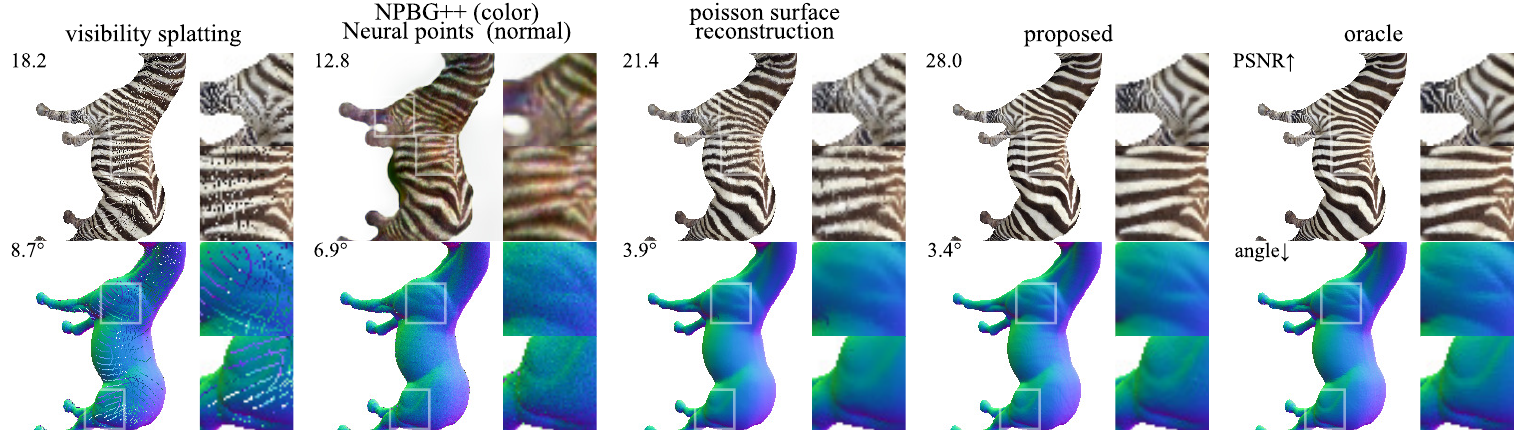}
	\end{subfigure}
	\vspace{-6mm}
	\caption{Example results of pointersect and baselines. Please see supplementary material for novel-view rendering videos. %
 }
	\label{fig: basic}
\end{figure*}

We use three datasets to evaluate the estimated intersection points, surface normal, and blending weights.
\vspace{-0.5em}
\begin{itemize}[leftmargin=*]
    \setlength\itemsep{-.1em}
	\item 7 meshes provided by \citet{Texturemontage05}, including the Stanford Bunny, Buddha, \etc.
	\item 30 meshes in ShapeNet Core dataset \cite{chang2015shapenet} containing sharp edges, including chairs, rifles, and airplanes.
	\item 13 test meshes in the sketchfab dataset \cite{qian2020pugeo}.
\end{itemize}
\vspace{-0.5em}
For each mesh, 6 RGBD images taken from front, back, left, right, top, and bottom are used to create the input point cloud; 144 output RGBD images at novel viewpoints are estimated and compared with the ground-truth rendering from Open3D.  See \autoref{fig: basic} for an illustration of the camera poses.
All cameras are $200 \times 200$ resolution and has a field-of-view of 30 degrees. 
The input RGBD images are used only to construct the input point cloud, \ie, we do not extract any image-patch features from the RGB images (except for NPBG++, see below).

We compare with four baselines: 
\vspace{-0.5em}
\begin{itemize}[leftmargin=*]
    \setlength\itemsep{-.1em}
	\item \textit{Visibility splatting} is the default point-cloud visualization method in Open3D. We set the point size to 1 pixel.  We use Open3D to estimate vertex normal at input points. The main purpose of including the baseline is to illustrate the input point cloud from the target viewpoint.
	\item \textit{Screened Poisson surface reconstruction} \cite{kazhdan2013screened} fits a scene-specific indicator function (1 inside the object and 0 outside) to reconstruct a surface from the point cloud. It is the workhorse for point-based graphics. We use the implementation in Open3D with the default parameters, and we provide ground-truth vertex normal from the mesh.
	\item \textit{NPBG++} \cite{rakhimov2022npbg++} uses rasterization and downsampling to fill in holes on the image plane. It supports unseen point clouds but requires input RGB images to extract features.  %
    We use the implementation and pretrained model from the authors.  We do not perform scene-specific finetuning.
	\item \textit{Neural Points} \cite{feng2022np} fits local implicit functions on the point clouds. Once the functions are fit, they can be used to increase the sampling rate and estimates surface normals. We use the original implementation and pretrained model from the authors. For visualization, we use the method to upsample the point clouds by 96 times. 
	\item \textit{Oracle} directly rasterizes the mesh. 
\end{itemize}
\vspace{-0.5em}
For pointersect, we use $k=40$, $\delta = 0.1$, and image-based rendering for all experiments unless otherwise noted.
We provide only point-wise $xyz$ and $rgb$ to pointersect; no other feature is used.
For all methods, we compute the Root Mean Square Error (RMSE) of the intersection point estimation, the average angle between the ground-truth and the estimated normal, the accuracy on whether a camera ray hit the surface, the Peak Signal-to-Noise Ratio (PSNR), Structural Similarity Index Measure (SSIM) \cite{wang2004image}, and LPIPS \cite{zhang2018perceptual} between ground-truth and estimated color images.
To ensure fair comparisons between all methods, we compute the errors of surface normal and depth map only for camera rays that both ground-truth and the testing method agree to hit a surface.

The results are shown in \autoref{table: basic}, and we provide examples in \autoref{fig: basic}.
Optimizing each input point cloud directly, Poisson surface reconstruction achieves high PSNR and low normal errors, outperforming the scene-agnostic baselines like NPBG++ and Neural Points.
Pointersect outperforms all prior methods even though it is scene-agnostic.
Trained on the sketchfab dataset, it performs best on the sketchfab dataset. 
Nevertheless, it still outperforms the scene-specific Poisson reconstruction on unseen meshes in the ShapeNet dataset and performs comparably on the tex-models dataset.
Moreover, pointersect is the only method supporting estimating intersection points, surface normal, and blending weights of any query ray without scene-specific optimization.

\begin{table}[t]
	\caption{Test results on unseen meshes in three datasets. }
	\label{table: basic}
    \vspace{-3mm}
	\centering
	\begin{adjustbox}{max width=\linewidth}
		\begin{tabular}{lllll}
			\toprule
			\makecell{Method} &
			\makecell{Metrics} & 
			\makecell{tex-models} & 
			\makecell{ShapeNet} &
			\makecell{Sketchfab} 
			\\
			\midrule
			
            \multirow{6}{4em}{Visibility splatting}  %
            & depth (RMSE) $\, \downarrow$ & $0.25 \pm 0.20$ & $0.08 \pm 0.08$ & $0.20 \pm 0.12$ \\
            & normal (angle (\degree)) $\, \downarrow$ & $12.57 \pm 4.60$ & $11.78 \pm 5.40$ & $14.04 \pm 4.73$ \\
            & hit (accuracy (\%)) $\, \uparrow$ & $98.0 \pm 1.5$ & $98.5 \pm 1.7$ & $98.0 \pm 2.0$ \\
            & color (PSNR (dB))$\, \uparrow$ & $19.2 \pm 2.2$ & $22.5 \pm 4.1$ & $19.6 \pm 2.9$ \\
            & color (SSIM)$\, \uparrow$ & $0.8 \pm 0.2$ & $0.9 \pm 0.1$ & $0.7 \pm 0.2$ \\
            & color (LPIPS)$\, \downarrow$ & $0.19 \pm 0.13$ & $0.10 \pm 0.10$ & $0.21 \pm 0.13$ \\
            \graymidrule
            \multirow{6}{4em}{Poisson surface recon.}  %
            & depth (RMSE) $\, \downarrow$ & $\mathbf{0.02 \pm 0.04}$ & $\mathbf{0.03 \pm 0.05}$ & $0.06 \pm 0.08$ \\
            & normal (angle (\degree)) $\, \downarrow$ & $8.48 \pm 3.97$ & $16.89 \pm 7.96$ & $14.58 \pm 6.29$ \\
            & hit (accuracy (\%)) $\, \uparrow$ & $\mathbf{99.8 \pm 0.1}$ & $98.8 \pm 1.8$ & $99.5 \pm 0.5$ \\
            & color (PSNR (dB))$\, \uparrow$ & $25.7 \pm 2.4$ & - & $25.0 \pm 3.1$ \\
            & color (SSIM)$\, \uparrow$ & $0.9 \pm 0.0$ & $0.9 \pm 0.1$ & $0.9 \pm 0.1$ \\
            & color (LPIPS)$\, \downarrow$ & $0.08 \pm 0.04$ & $0.06 \pm 0.05$ & $0.10 \pm 0.04$ \\
            \graymidrule
            \multirow{6}{4em}{Neural points \cite{feng2022np}}  %
            & depth (RMSE) $\, \downarrow$ & $0.07 \pm 0.10$ & $0.04 \pm 0.03$ & $0.06 \pm 0.05$ \\
            & normal (angle (\degree)) $\, \downarrow$ & $11.28 \pm 3.30$ & $17.14 \pm 6.01$ & $14.28 \pm 3.44$ \\
            & hit (accuracy (\%)) $\, \uparrow$ & $98.9 \pm 0.4$ & $98.9 \pm 0.7$ & $99.1 \pm 0.2$ \\
            & color (PSNR (dB))$\, \uparrow$ & not supp. & not supp. & not supp. \\
            & color (SSIM)$\, \uparrow$ & not supp. & not supp. & not supp. \\
            & color (LPIPS)$\, \downarrow$ & not supp. & not supp. & not supp. \\
            \graymidrule
            \multirow{6}{4em}{NPBG++ \cite{rakhimov2022npbg++}}  %
            & depth (RMSE) $\, \downarrow$ & not supp. & not supp. & not supp. \\
            & normal (angle (\degree)) $\, \downarrow$ & not supp. & not supp. & not supp. \\
            & hit (accuracy (\%)) $\, \uparrow$ & not supp. & not supp. & not supp. \\
            & color (PSNR (dB))$\, \uparrow$ & $16.5 \pm 2.1$ & $19.3 \pm 4.0$ & $18.0 \pm 1.6$ \\
            & color (SSIM)$\, \uparrow$ & $0.7 \pm 0.1$ & $0.8 \pm 0.1$ & $0.7 \pm 0.1$ \\
            & color (LPIPS)$\, \downarrow$ & $0.27 \pm 0.07$ & $0.18 \pm 0.08$ & $0.24 \pm 0.08$ \\
            \graymidrule
            \multirow{6}{4em}{Proposed}  %
            & depth (RMSE) $\, \downarrow$ & $0.05 \pm 0.09$ & $\mathbf{0.03 \pm 0.03}$ & $\mathbf{0.05 \pm 0.04}$ \\
            & normal (angle (\degree)) $\, \downarrow$ & $\mathbf{6.77 \pm 2.71}$ & $\mathbf{11.29 \pm 5.08}$ & $\mathbf{8.53 \pm 2.79}$ \\
            & hit (accuracy (\%)) $\, \uparrow$ & $\mathbf{99.8 \pm 0.2}$ & $\mathbf{99.6 \pm 0.6}$ & $\mathbf{99.8 \pm 0.1}$ \\
            & color (PSNR (dB))$\, \uparrow$ & $\mathbf{28.2 \pm 1.9}$ & $\mathbf{28.0 \pm 6.4}$ & $\mathbf{28.1 \pm 2.7}$ \\
            & color (SSIM)$\, \uparrow$ & $\mathbf{1.0 \pm 0.0}$ & $\mathbf{1.0 \pm 0.0}$ & $\mathbf{0.9 \pm 0.0}$ \\
            & color (LPIPS)$\, \downarrow$ & $\mathbf{0.04 \pm 0.03}$ & $\mathbf{0.04 \pm 0.04}$ & $\mathbf{0.06 \pm 0.04}$ \\
            
			\bottomrule
		\end{tabular}
	\end{adjustbox}
\end{table}

\begin{table}[t]
	\caption{Test results on the Hypersim dataset. The test is conducted on 10 new images not used to create the point cloud.}
	\label{table: hypersim}
	\vspace{-3mm}
	\centering
	\begin{adjustbox}{max width=\linewidth}
		\begin{tabular}{llllll}
			\toprule
			\makecell{Metric} &
			\makecell{Vis. splatting} & 
			\makecell{Poisson recon.} & 
			\makecell{NPBG++} &
            \makecell{NGP \cite{mueller2022instant}} &
			\makecell{Proposed} 
			\\
			\midrule
			PSNR (dB) $\uparrow$ & 
			$11.7 \pm 1.1$ &
			$26.3 \pm 3.1$ &
			$27.9 \pm 3.7$ &
            $28.5 \pm 5.8$ &
			$\mathbf{29.8 \pm 5.1}$ \\
            SSIM $\uparrow$ & 
			$0.12 \pm 0.05$ &
			$0.85 \pm 0.06$ &
			$0.89 \pm 0.04$ &
            $0.90 \pm 0.06$ &
			$\mathbf{0.91 \pm 0.04}$ \\
            LPIPS $\downarrow$ & 
			$0.79 \pm 0.04$ &
			$0.37 \pm 0.04$ &
			$0.33 \pm 0.03$ &
            $0.31 \pm 0.08$ &
			$\mathbf{0.25 \pm 0.04}$ \\
			\bottomrule
		\end{tabular}
	\end{adjustbox}
\end{table}

\subsection{Room-scale rendering with RGBD images}
\label{sec: exp hypersim}

Next, we evaluate pointersect on a room-scale scene, whose geometry is very different from the training meshes.
We randomly select a room scene in the Hypersim dataset~\cite{roberts2021hypersim} which contains 100 RGBD images captured in the room.
We randomly select 90 of them to construct the input point cloud and use the rest for evaluation.
We downsample the images by 2 (from $1024 \times 768$ to $512 \times 384$), and we use uniform voxel downsampling to reduce the number of points (the room size is 4 units and the voxel size is 0.01 units). 
We provide the ground-truth vertex normal from the dataset to Poisson reconstruction.
We also train a state-of-the-art NeRF method, NGP \cite{mueller2022instant, KaolinWispLibrary}, using the same 90 input images, for 1000 epochs, taking 2 hours on 1 A100 GPU.  %
Note that NGP and pointersect are not directly comparable---NGP utilizes scene-specific optimization whereas pointersect utilizes depth.
We provide it as a reference baseline.

\begin{figure}[t]
	\centering
	\vspace{-3mm}
	\hspace{-1mm}
	\includegraphics[width=0.9\linewidth]{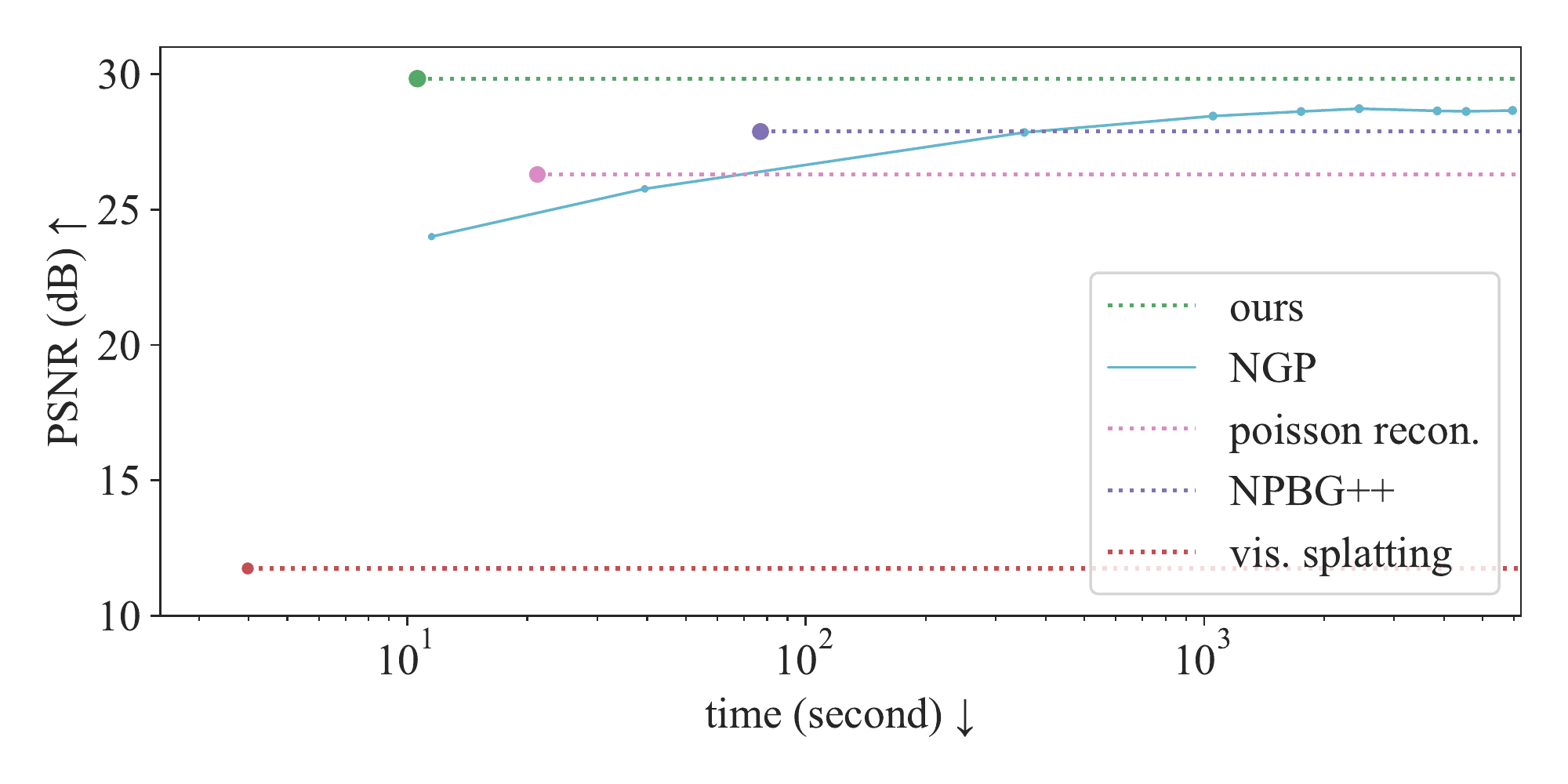}
	\vspace{-3.5mm}
	\caption{PSNR \vs time of \autoref{fig: hypersim}.  The x axis is the time to render 10 validation images of resolution $512 \times 384$.  We train the NGP \cite{mueller2022instant, KaolinWispLibrary} from 1 to 1000 epochs, and the time includes both training on the 90 input images and the rendering.  %
	}
	\label{fig: hypersim psnr time}
\end{figure}

\begin{figure*}[t]
	\centering
    \vspace{-3mm}
	\includegraphics[width=\linewidth]{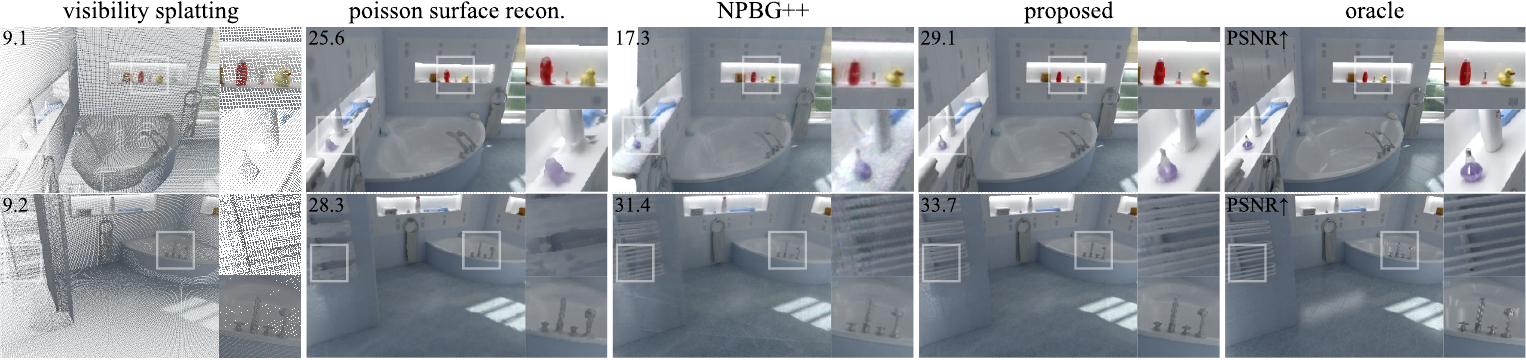}	
    \vspace{-6mm}
	\caption{Room-scale point cloud rendering results.  Please see supplementary material for novel-view rendering videos. }
	\label{fig: hypersim}
\end{figure*}

\autoref{fig: hypersim psnr time}, \autoref{fig: hypersim}, and \autoref{table: hypersim} show the results on the 10 test RGBD images.
As can be seen, while pointersect is trained on small meshes, it generalizes to room-scale scenes and outperforms prior state-of-the-art baselines.

\begin{figure*}[t]
	\centering
	\vspace{-3mm}
	\includegraphics[width=\linewidth]{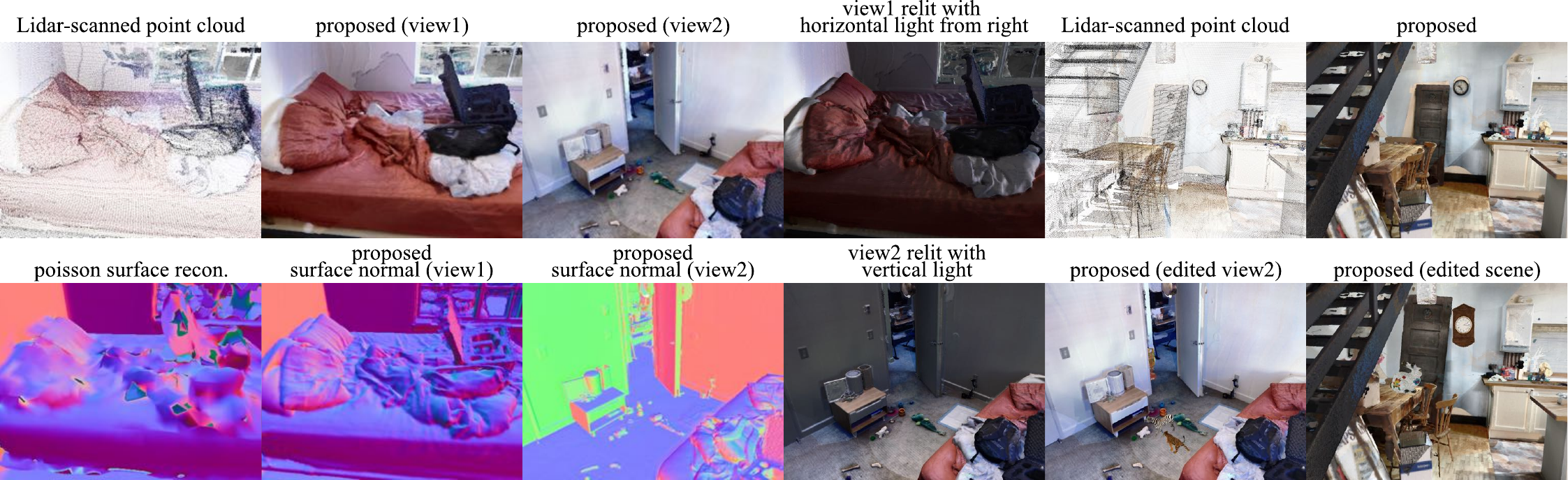}	
	\vspace{-6mm}
	\caption{We test our model on a Lidar-scanned point cloud. Due to the small amount of noise, the quality of Poisson reconstruction degrades significantly. In comparison, pointersect is less affected. With the estimated surface normal, we re-render the scene with directional-dominant light using the colors as albedo. Using point cloud as the scene representation enables easy scene editing. We edit the input point clouds and render the new scene using pointersect. Can you spot the differences? Please see supplementary material for novel-view rendering, relighting, and scene-editing videos. \ccLogo~clock: \citet{clock}.}
	\label{fig: arkitscenes}
\end{figure*}

\subsection{Ray tracing with point clouds}
\label{sec: exp ray trace}

As mentioned before in \Cref{sec: pointersect}, pointersect provides a unique capability of ray-tracing with global illumination that is difficult to achieve with methods like NeRFs. 
We build a simple path tracer in PyTorch, following \cite{shirley2018ray}.
We also implement the Cook-Torrance microfacet specular shading model (with the split sum approximation), following \cite{karis2013real}.
We construct a scene composed of a Lambertian floor (albedo = 0.2), a vertical mirror (kbase = 0.5, roughness = 0, metallic = 1, specular = 0.5), a sphere (kbase = 0.7, roughness = 0.7, metallic = 0.5, specular = 0), and an all-white environment map.
The floor and the mirror are represented analytically, and the sphere is represented with 5000 points. %
We trace 4 bounces and 2000 rays per pixel. 

As shown in \autoref{fig:teaser}b, the result clearly shows the effect of global illumination, \eg, the reflection of the sphere in the mirror and the cast shadow on the floor.  %

\subsection{Inverse rendering}
\label{sec: exp inverse}

Another unique capability of pointersect is its differentiability.
As a neural net, pointersect allows gradient calculation of color (blending weights) and surface normal with respect to the point cloud (\eg, $xyz$ and $rgb$).
We optimize a noisy point cloud's $xyz$ and $rgb$ given 100 clean input RGB images, camera poses, and binary foreground segmentation maps.
The results (\autoref{fig: inv detail} and \autoref{table: inv} in the supplementary) show that the optimization effectively denoises the point cloud.
Please refer to \autoref{sec: inverse rendering details} for details.

\begin{table}[t]
	\caption{Optimize a noisy point cloud with pointersect.}
	\label{table: inv}
	\vspace{-3mm}
	\centering
	\begin{adjustbox}{max width=\linewidth}
		\begin{tabular}{lcccc}
			\toprule
			& %
			\multicolumn{2}{c}{100 input views} & 
			\multicolumn{2}{c}{144 novel views} 
			\\
			& 
			before opt.& 
			after opt. & 
			before opt. & 
			after opt. 
			\\
			\midrule
			PSNR (dB) $\uparrow$ & 
			$10.1 \pm 0.5$&   %
			$24.6 \pm 0.5$ &  %
			$13.9 \pm 0.8$ &   %
			$25.7 \pm 1.2$   %
			\\  
			normal (angle (\degree)) $\, \downarrow$ & 
			$54.0 \pm 0.9$&   %
			$17.8 \pm 2.1$ &  %
			$55.3 \pm 0.4$ &   %
			$18.3 \pm 2.7$    %
			\\  
			depth (rmse) $\, \downarrow$ & 
			$0.46 \pm 0.08$&   %
			$0.09 \pm 0.06$ &  %
			$0.33 \pm 0.08$ &   %
			$0.10 \pm 0.07$   %
			\\  
			\bottomrule
		\end{tabular}
	\end{adjustbox}
	 \vspace{-2mm}
\end{table}

\subsection{Real Lidar-scanned point clouds}
\label{sec: exp real}

Finally, we test pointersect on real Lidar-scanned point clouds. 
The goal is to evaluate how it handles a small amount of noise in scanned point clouds, even though it is trained on clean ones.
We take two Lidar point clouds from the ARKitScenes dataset \cite{dehghan2021arkitscenes}, perform uniform voxel downsampling to reduce the number of points (the voxel size is $0.01$ and the scene size is around $14$ units), and rendered with Poisson reconstruction and pointersect (with $k = 100$ and $\delta=0.2$ since the scene is larger and not unit-length normalized).
As can be seen in \autoref{fig: arkitscenes}, our model successfully renders the real point clouds, whereas the output quality of Poisson reconstruction degrades significantly.

\paragraph{Scene editing.}
One advantage of utilizing point clouds as the scene representation instead of an implicit representation like NeRF is that we can easily edit the scene (by directly moving, adding, removing points).
In \autoref{fig: arkitscenes}, we present results of scene relighting which utilizes the estimated surface normal and scene editing where we insert new and change the size and location of point-cloud objects in the scene.

\section{Discussions}

\paragraph{Rendering speed.}
Pointersect requires one transformer evaluation per query ray. 
In contrast, NeRF and SDF methods require multiple evaluations per ray (in addition to per-scene training).
\autoref{fig: hypersim psnr time} shows the time to render 10 test images in \Cref{sec: exp hypersim}.
With a resolution of $512 \times 384$ and 700k points, the current rendering speed of pointersect is $\sim1$ frame per second (fps) with unoptimized Python code.
The speed can be greatly improved with streamlined implementation and accelerated attention  \cite{kitaev2020reformer, marin2021token}.
See \autoref{sec: runtime} for detailed complexity and runtime analysis.

\paragraph{Known artifacts.}
Currently, our pointersect model produces two types of artifacts.
First, pointersect may generate floating points when a query ray is near an edge where the occluded background is far away 
 or at the middle of two parallel edges.
Second, we currently pass a fixed number of neighboring points to the pointersect model. Thus when there are multiple layers of surfaces, the actual (\ie, first) surface may receive only a small number of points, reducing the output quality.

\paragraph{Connection to directed distance fields.}
\citet{aumentado2022representing} and \citet{feng2022prif} %
propose to represent a mesh with a Directed Distance Field (DDF).  %
Similar to a SDF, a DDF, $\mD(\r)$, is learned specifically for each mesh, and it records the traveling distance of ray $\r$ to the nearest surface.
The proposed pointersect, $f(\r, \cP)$, can be thought of as an estimator of DDFs.  %
\citet{aumentado2022representing} derive several geometric properties of DDFs, which also apply to pointersect.

\section{Conclusion}

We have introduced pointersect, a novel method to render point clouds as if they are surfaces.
Compared to other scene representations like implicit functions and prior point-cloud rendering methods, pointersect provides a unique combination of capabilities, including direct rendering of input point clouds without per-scene optimization, direct surface normal estimation, differentiability, ray tracing with global illumination, and intuitive scene editing.  
With the ubiquitousness of point clouds in 3D capture and reconstruction, we believe that pointersect will spur innovations in computer vision as well as virtual and augmented reality technology.

\paragraph{Acknowledgment.}
We thank Yi Hua for all the interesting general discussions about geometry.

{\small
\bibliographystyle{ieee_fullname_natbib}

}

\pagebreak
\appendix

In the supplementary material, we provide details about the following topics:
\vspace{-0.5em}
\begin{itemize}[leftmargin=*]
	\setlength\itemsep{-.1em}
	\item \textit{overview of related work} in \Cref{sec: additional related work};
	\item \textit{accelerated structure}  in \Cref{sec: pr detail};
	\item \textit{model architecture} in \Cref{sec: architecture details};
	\item \textit{complexity and runtime analyses} in \Cref{sec: runtime};
	\item \textit{training procedure} in \Cref{sec: model details}; 
	\item \textit{inverse rendering} in \Cref{sec: inverse rendering details};
	\item \textit{Chamfer distance} in \Cref{sec: chamfer};
	\item \textit {additional results} to evaluate and gain insights on pointersect, including:
	\vspace{-0.5em}
	\begin{itemize}[leftmargin=*]
		\setlength\itemsep{-.1em}
		\item novel-view rendering video and additional scenes in \Cref{fig: additional basic} and in the offline web page, 
		\item ablation study on the number of input views and the choice of $k$ in \Cref{sec: num views and k},
		\item ablation study on the input resolution in \Cref{sec: num views frontal}, 
		\item and the effect of ground-truth vertex normal to Poisson reconstruction in \Cref{sec: gt vertex normal};
	\end{itemize}
	\item \textit{Effect of noise in depth map} in \Cref{sec: noisy arkit};
	\item and finally, the \textit{entire training dataset} containing 48 meshes and their credits in \Cref{sec: sketchfab dataset}.
	\vspace{-0.5em}
\end{itemize}

\section{Additional related work}
\label{sec: additional related work}

In this paper, we focus on comparing with point-cloud rendering methods that do not require per-scene optimization. 
In this section, we briefly discuss and include additional related work. 
We also provide an overview in \cref{table: related works} for interested readers.

Many recent works develop novel view synthesis techniques given only the RGB images and optionally their camera information.
Neural Radiance Field (NeRF)~\cite{mildenhall2021nerf} achieve a great success by representing the volume density and the radiance field with a neural network, and it is able to render high-quality photo-realistic images from novel viewpoints.
Many follow-up works further improve the quality~\cite{barron2021mip,verbin2022refnerf}, 
the training efficiency~\cite{reiser2021kilonerf, yu2021plenoctrees}, 
the rendering speed ~\cite{muller2022instant,fridovich2022plenoxels}, 
and the generalization capability~\cite{yu2021pixelnerf,wang2021ibrnet}.
Additional capabilities have also been introduced into NeRF, including estimating shape and reflectance~\cite{boss2021nerd, srinivasan2021nerv,zhang2021nerfactor}.
Several methods also aim to reduce or completely avoid the per-scene optimization of NeRF by utlizing image features~\cite{yu2021pixelnerf, wang2021ibrnet,chen2021mvsnerf}.
These methods often rely on additional information, including class labels,

Other rendering primitives, including spheres~\cite{lassner2021pulsar}, occupancy field~\cite{niemeyer2020differentiable}, Signed Distance Function (SDF)~\cite{yariv2020multiview,kellnhofer2021neural}, light field~
\cite{sitzmann2021light,suhail2022light}, or a plane sweep volume~\cite{choi2019extreme,chen2021mvsnerf,mildenhall2019local} have also been developed. 

Hybrid methods have also been developed.  
These methods first utilize Structure from Motion (SfM) or multi-view geometry to estimate scene geometry form the input RGB images.
The estimated geometry is then used to provide additional supervision for NeRF~\cite{deng2022depth,roessle2022dense} or to anchor feature aggregation~\cite{riegler2020free,riegler2021stable}.

As mentioned earlier, this paper focuses on point-cloud rendering \textit{without} per-scene optimization. 
The methods discussed above operate in a different setting or require per-scene training.
Therefore, while we are inspired by many of these methods, our method is not directly comparable.

\begin{table*}[t]
\caption{\textbf{Rendering methods for 2D images and point clouds.} The table provides a summary of the scene presentation (geometry primitives) and the capabilities of various methods. 
\cmarkstar: require vertex normal as inputs. 
\cmarkflat: provides good results, and per-scene optimization further improves the quality.
\hmark: depth and normal can be estimated from the density function.
}
\label{table: related works}
\centering
\begin{adjustbox}{max width=\linewidth}
\begin{tabular}{lllcccc}
\toprule
Category & Method & Primitives & \makecell{Render \\ color} & \makecell{Estimate \\ depth }& \makecell{Estimate \\ normal} & \makecell{No per-scene \\ optimization} \\
\midrule 
\multirow{19}{*}{Rendering from 2D images} & NeRF~\cite{mildenhall2021nerf} & volume density (MLP) & \cmark & \hmark & \hmark & \xmark \\
 & NeRD~\cite{boss2021nerd} & volume density (MLP) & \cmark & \cmark & \cmark & \xmark \\
 & NeRV~\cite{srinivasan2021nerv} & volume density (MLP) & \cmark & \cmark & \cmark & \xmark \\
 & NeRFactor~\cite{zhang2021nerfactor} & volume density (MLP) & \cmark & \cmark & \cmark & \xmark \\
 & KiloNeRF~\cite{reiser2021kilonerf} & volume density (grid MLP) & \cmark & \hmark & \hmark & \xmark \\
 & InstantNGP~\cite{muller2022instant} & volume density (multi-resolution MLP) & \cmark & \hmark & \hmark & \xmark \\
 & Plenoctrees~\cite{yu2021plenoctrees} & volume density (octree) & \cmark & \hmark & \hmark & \xmark \\
 & Plenoxels~\cite{fridovich2022plenoxels} & volume density (voxel) & \cmark & \hmark & \hmark & \xmark \\
 & PixelNeRF~\cite{yu2021pixelnerf} & volume density (MLP + 2D features) & \cmark & \hmark & \hmark & \cmark \\
 & IBRNet~\cite{wang2021ibrnet} & volume density (MLP + 2D features) & \cmark & \hmark & \hmark & \cmarkflat \\
  & Pulsar~\cite{lassner2021pulsar} & spheres & \cmark & \cmark & \cmark & \xmark \\
 & Differentiable Volumetric Rendering~\cite{niemeyer2020differentiable} & occupancy field & \cmark & \cmark & \cmark & \xmark \\
 & IDR~\cite{yariv2020multiview} & SDF & \cmark & \cmark & \cmark & \xmark \\
 & Neural Lumigraph Rendering~\cite{kellnhofer2021neural} & SDF & \cmark & \cmark & \cmark & \xmark \\
 & Light field Networks~\cite{sitzmann2021light} & light field & \cmark & \xmark & \xmark & \xmark \\
 & Light Field Neural Rendering~\cite{suhail2022light} & light field & \cmark & \xmark & \xmark & \xmark \\
 & Extreme View Synthesis~\cite{choi2019extreme} & plane sweep volume & \cmark & \cmark & \xmark & \cmark \\
 & MVSNeRF~\cite{chen2021mvsnerf} & plane sweep volume & \cmark & \cmark & \hmark & \cmarkflat \\
 & Local Light Field Fusion~\cite{mildenhall2019local} & plane sweep volume + light field & \cmark & \cmark & \xmark & \cmark \\
 \graymidrule 
\multirow{4}{*}{Rendering from 2D images + SfM Depth} & Depth-supervised NeRF~\cite{deng2022depth} & volume density (MLP) & \cmark & \cmark & \hmark & \xmark \\
 & Dense Depth Priors~\cite{roessle2022dense} & volume density (MLP) & \cmark & \cmark & \hmark & \xmark \\
 & Free View Synthesis~\cite{riegler2020free} & mesh & \cmark & \cmark & \cmark & \cmark \\
 & Stable View Synthesis~\cite{riegler2021stable} & mesh & \cmark & \cmark & \cmark & \cmark \\  
 \midrule 
\multirow{5}{*}{\makecell[l]{Point cloud to \\ other representations}} & Poisson reconstruction~\cite{kazhdan2006poisson} & indicator func. & \cmark & \cmark & \cmarkstar & \xmark \\ %
 & Shape as point~\cite{peng2021shape} & indicator func. & \xmark & \cmark & \cmarkstar & \xmark \\
 & Neural pull~\cite{ma2021neural} & SDF & \xmark & \cmark & \cmark & \xmark \\
 & Point2Mesh~\cite{hanocka2020point2mesh} & mesh & \xmark & \cmark & \cmark & \xmark \\
 & Neural point~\cite{feng2022np} & neural point & \xmark & \cmark & \cmark & \cmark \\
\graymidrule 
\multirow{5}{*}{\makecell[l]{Point cloud rasterization}} & Visibility Splatting~\cite{pfister2000surfels} & surfel & \cmark & \cmark & \cmarkstar & \cmark \\
 & NPBG~\cite{aliev2020neural} & points & \cmark & \xmark & \xmark & \xmark \\
 & ADOP~\cite{ruckert2022adop} & points & \cmark & \xmark & \xmark & \xmark \\
 & Multi-plane~\cite{dai2020neural} & points & \cmark & \xmark & \xmark & \xmark \\
 & NPBG++~\cite{rakhimov2022npbg++} & points & \cmark & \xmark & \xmark & \cmark \\
 \graymidrule
\multirow{4}{*}{\makecell[l]{Point cloud ray-casting}} & Iterative ray-surface intersection~\cite{adamson2003approximating} & points & \cmark & \cmark & \cmark & \xmark \\
 & Point-Nerf~\cite{xu2022point} & points & \cmark & \cmark & \hmark & \cmarkflat \\
 & NPLF~\cite{ost2022neural} & points & \cmark & \xmark & \xmark & \xmark \\
 & Ours (pointersect) & points & \cmark & \cmark & \cmark & \cmark \\
\bottomrule
\end{tabular}
\end{adjustbox}
\end{table*}

\section{Finding points along a ray}
\label{sec: pr detail}

Finding the intersection point of a ray $\r = (r_o, \vec{r_d})$ with a point cloud $\cP$ requires only points near the ray.
Thus, we pass only the $k$ nearest points within a cylinder of radius $\delta$ in terms of their perpendicular distances to the ray.
A naive implementation would examine all points in $\cP$, sort them according to their distances, and keep the $k$ nearest ones, taking $\mO(n \log n)$ operations per ray, where $n$ is the total number of points in $\cP$.
Since we only need points with the cylinder, a commonly used strategy to reduce the time complexity is to build an accelerated structure like octree to reduce the number of candidate points to examine \cite{pharr2016physically}.
Building an octree takes $\mO(n \log n)$ operations, and searching for nearby points takes $\mO(\log n)$ operations per ray.
For a static scene, the octree can be built once and keep reusing the tree to find neighbor points.

\begin{figure}[t]
	\centering
	\includegraphics[width=\linewidth]{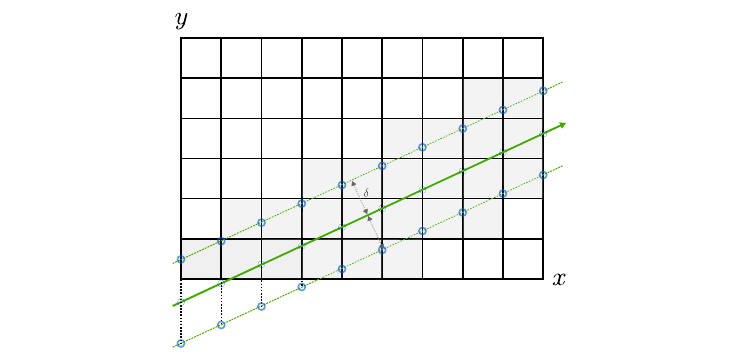}	
	\vspace{-6mm}
	\caption{The voxel grid structure utilizes the easy computation of grid-ray intersection. Note that we need to trace the ray along the slowest moving axis (in this case the $x=c_i$ planes). } 
	\label{fig: pr}
\end{figure}

While octree greatly improves the speed for static $\cP$, for dynamic scenes or for inverse rendering, where many points in $\cP$ can change at every iteration, rebuilding the tree at every iteration becomes an overhead.
We build an accelerated structure that can be built in parallel on GPU to improve the speed of inverse rendering.
The structure is based on a voxel grid.
Intuitively, as can be seen from \figref{fig: pr}, we divide the space into voxels, and we record the points contained in each voxel with a table.
Thus, building the voxel grid structure takes $\mO(n)$ operations, where $n$ is the total number of points.
When a ray is given, we trace the ray through the voxel grid. 
The grid-ray intersection can be computed in $\mO(g)$ time for a given ray, where $g$ is the number of grid cells per dimension, and the computation for each ray can be computed in parallel using multiple GPU threads. 

Overall, the time complexity of finding $k$ nearest points within a surrounding cylinder of radius $\delta$ is $\mO\left(n + mg + m q \log (q) \right)$, where $q$ is the number of points lying within the cylinder, and the last term corresponds to the cost of sorting the distances from the points to the ray.  In the case, when the points are uniformly distributed, in expectation $q = n/g^3$
Note that our implementation is not optimized; for example, the optimal time complexity of retrieving an unordered $k$ nearest points from the $q$ points is $\mO(q)$. 
This structure can be slower than using an octree for static scenes, which takes $\n \log(n) + m \log(n) + m q$, but its faster construction makes it more suitable for dynamic point clouds.
We also want to point out that while we implement the accelerated structure in CUDA, the implementation is not optimized, and the speed can be further improved, as we will see in \cref{sec: runtime}.

\section{Architecture details}
\label{sec: architecture details}

\begin{figure}[t]
	\centering
    \includegraphics[width=\linewidth]{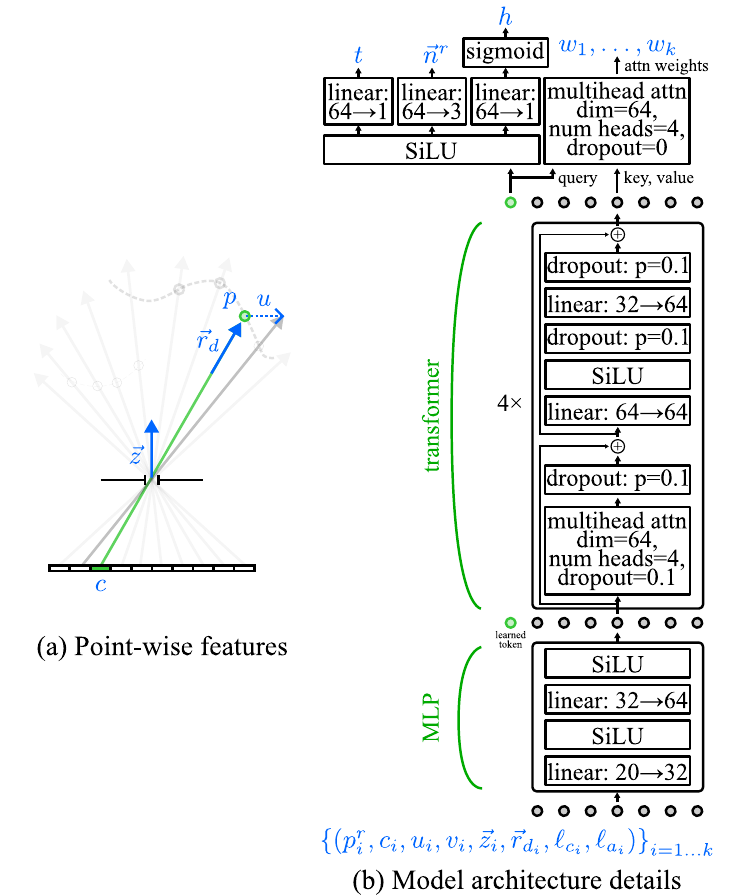}
	\caption{(a) Point-wise features extracted from camera rays and (b) model architecture details.}
	\label{fig: architecture details}
\end{figure}

We use a transformer to build the pointersect model. 
The model architecture details, including the number of layers and the layer composition, are illustrated in \cref{fig: architecture details}b.

The model architecture is composed of a Multi-layer Perceptron (MLP) and a transformer. 
The MLP is used to convert the input features (\eg, $xyz$, $rgb$) into the dimensionality used by the transformer (which is 64). 
We do not use positional encoding for $xyz$ like NeRF \cite{mildenhall2021nerf}---we found that adding positional encoding reduces the estimation accuracy.

We use the standard transformer block from \citet{vaswani2017attention}.
We remove the layer normalization layer, and we use a dropout probability of $0.1$.
The transformer is composed of 4 layers of transformer blocks and has a dimension of 64.
We additional learn a token and insert it at the input of the transformer, in order to estimate the ray traveling distance, the surface normal, and the probability of hitting a surface.
We use the SiLU nonlinearity \cite{ramachandran2017searching,hendrycks2016gaussian}.

The output of the transformer contain $k+1$ tokens---1 corresponding to the special token and $k$ for the neighboring points.
We use a linear layer to convert the special token to the corresponding estimates.
To compute the material blending weights, $\w = \left[w_1, \dots, w_k \, | \, w_i \in [0, 1], \sum_{i=1}^k w_i = 1 \right]$, we use a multihead attention layer and take its output attention weights as $\w$. 
In other words, the tokens are projected by a linear layers as queries, keys, and values, and the softmax attention is used to compute the similarities between the tokens and the special token.

\section{Complexity analysis}
\label{sec: runtime}

\begin{figure*}[t]
	\centering
	\begin{subfigure}[t]{\linewidth}
		\captionsetup{justification=centering}
		\centering
		\begin{subfigure}[t]{0.245\linewidth}
			\captionsetup{justification=centering}
			\centering
			\includegraphics[width=\linewidth]{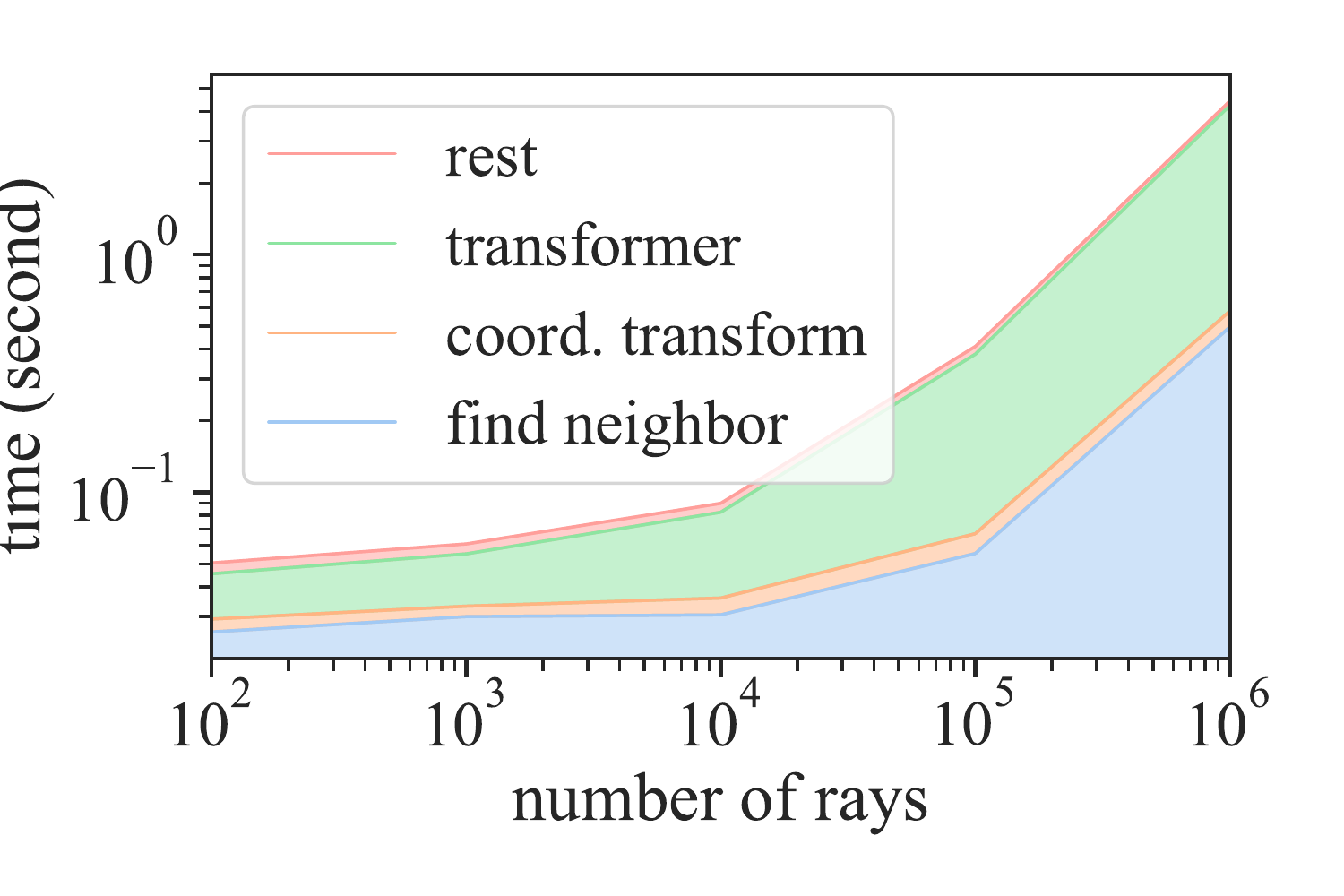}
			\vspace{-5mm}
			\caption{effect of number of rays. \\ number of points = 10,000\\$k = 40$}
		\end{subfigure}
		\hfill
		\begin{subfigure}[t]{0.245\linewidth}
			\captionsetup{justification=centering}
			\centering
			\includegraphics[width=\linewidth]{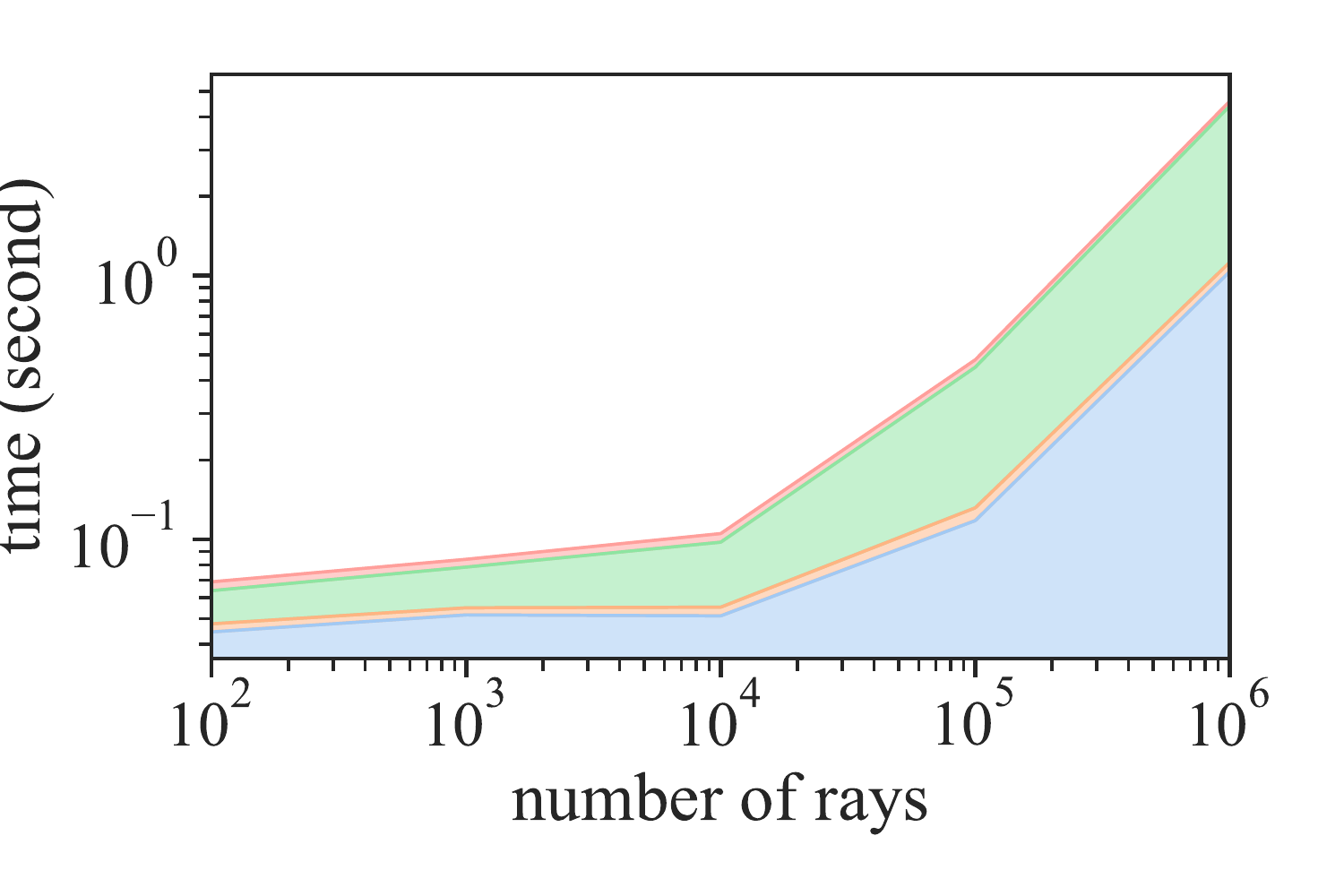}
			\vspace{-5mm}
			\caption{effect of number of rays. \\ number of points = 100,000 \\$k = 40$}
		\end{subfigure}
		\hfill
		\begin{subfigure}[t]{0.245\linewidth}
			\captionsetup{justification=centering}
			\centering
			\includegraphics[width=\linewidth]{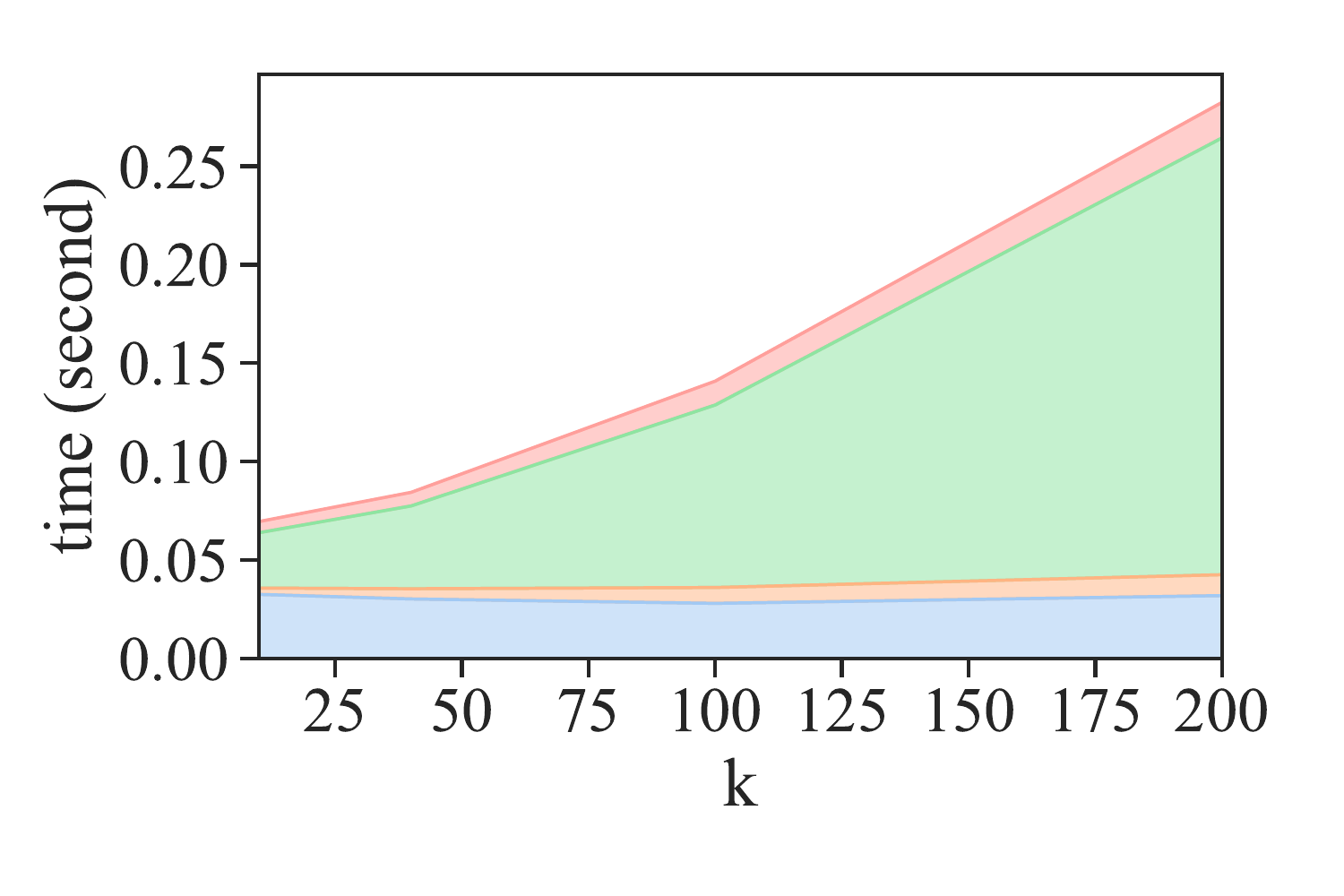}
			\vspace{-5mm}
			\caption{effect of $k$. \\ number of points = 10,000 \\ number of rays = 10,000 }
		\end{subfigure}
		\hfill
		\begin{subfigure}[t]{0.245\linewidth}
			\captionsetup{justification=centering}
			\centering
			\includegraphics[width=\linewidth]{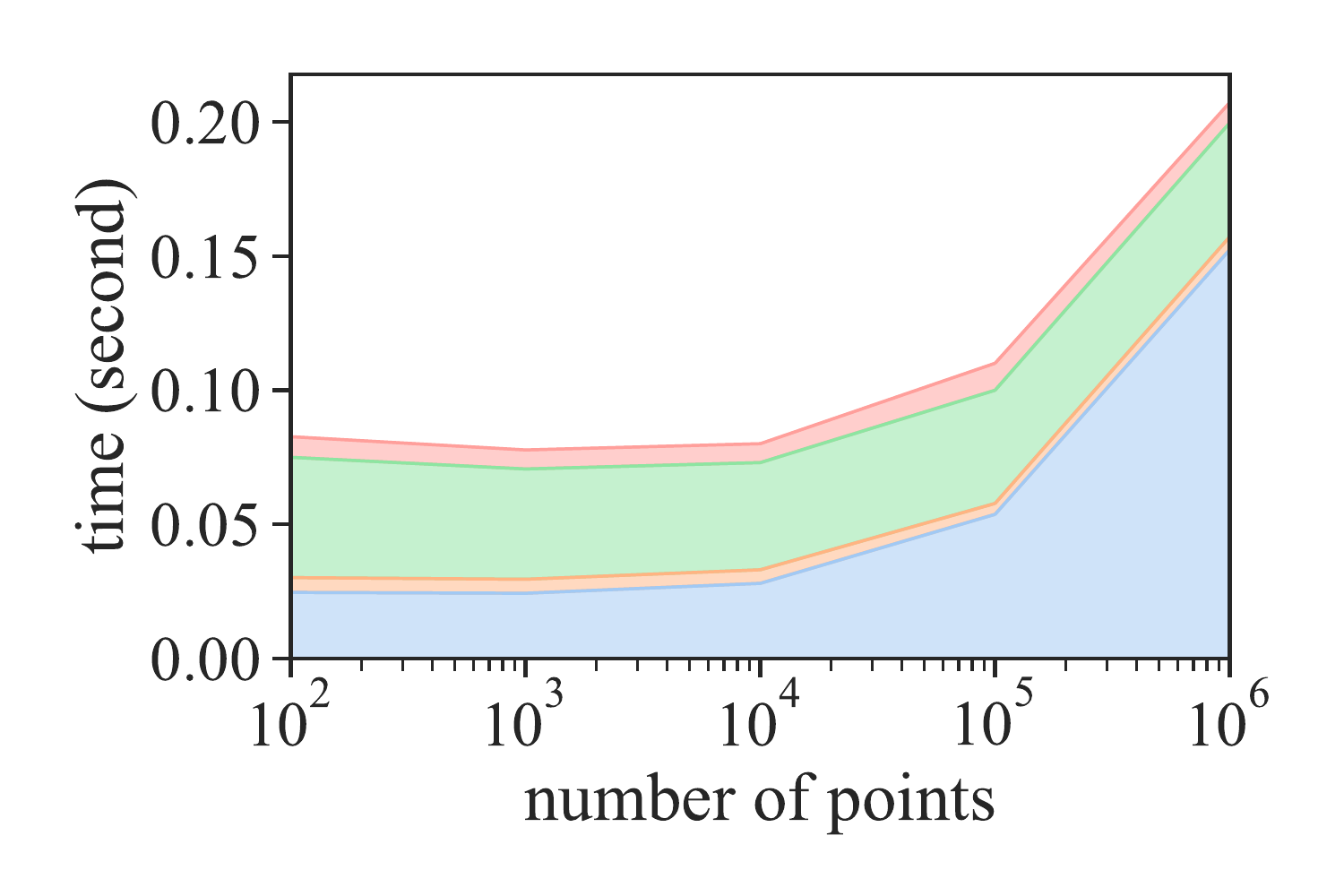}
			\vspace{-5mm}
			\caption{effect of number of points. \\ number of rays = 10,000 \\ $k = 40$}
		\end{subfigure}
	\end{subfigure}
	\vspace{-1mm}
	\caption{
		Runtime measurements of the proposed method rendering (a, b) different numbers of rays, (c) different numbers of neighboring points $k$, and (d) different number of points.  We break the total runtime into four main categories and illustrate them as cumulative plots.   In the experiment, we uniformly sample points within a square cube between $[-1, 1]^3$, uniformly sample ray origin in the cube, and uniformly sample rays toward all directions.  We repeat each case by three times and report the average.   The experiment is conducted on a single A100 GPU and PyTorch 1.10.1 with the typical implementation of transformer. 
	}
	\label{fig: runtime}
\end{figure*}

In this section, we analyze the computation complexity of our implementation. 
Pointersect is composed of three main operations:
\begin{itemize}[leftmargin=*]
	\setlength\itemsep{-.1em}
	\item finding points along the query ray (\Cref{sec: pr detail});
	\item transform to the canonical coordinate (\Cref{sec: pointersect});
	\item and run the transformer model (\Cref{sec: architecture details}).
\end{itemize}
Finding neighboring points, as discussed in \cref{sec: pr detail}, has a time complexity of $\mO(n + mg + m q \log(q) )$, where $n$ is the number of points, $m$ is the number of query rays, $g$ is the number of cells in each dimension of the grid, and $q$ is the total number of points lying within a cylinder of radius $\delta$ centered on the query ray.  In the case, when the points are uniformly distributed, in expectation $q = n/g^3$
The transformation to the canonical coordinate is $\mO(mk)$, where $k$ is the number of the nearest neighbor points. 
The transformer has a time complexity of $\mO(Lkd(k + d))$, where $L$ and $d$ are the number of layers and the dimension of the transformer, respectively.

\cref{fig: runtime} shows the runtime measurements of various settings. 
In the experiment, we simulate an adversarial scenario where the points are uniformly sampled within a square cube between $[-1, 1]^3$ (instead of on surfaces as typical scenarios).
As can be seen from the results, the main bottlenecks are finding neighboring points of a ray and the transformer.
While we implement the neighbor-point searching with a custom CUDA kernel, there is still a significant room for improvement.
Nevertheless, with a point cloud containing $n = 10,000$ points and $k = 40$, our non-optimized implementation is capable of rendering $100 \times 100$ images in 10 fps, $200 \times 200$ images in 5 fps, and $500 \times 400$ in 1 fps.

\section{Model training details}
\label{sec: model details}

We train the pointersect model with the 48 training meshes in the sketchfab dataset \cite{qian2020pugeo}.
We center and scale the meshes such that the longest side of their bounding box is 2 units in length. 
For each training iteration, we randomly select one mesh and randomly construct 30 input cameras and 1 target camera, which capture RGBD images using the mesh-ray intersection method in Open3D \cite{Zhou2018}.
We do not apply anti-aliasing filters on the RGBD images. 
This allows us to get the ground truth of the specific intersection point, instead of a blurred and average one across a local neighborhood.
The ground-truth RGBD images are rendered without global illumination.
This is intentional and allows the learned $\w$ to focus on material properties and not be affected by lighting conditions and cast shadows. 
Both input and target cameras have a field-of-view of 60 degrees.

We create the input point cloud using the input RGBD images. 
Specifically, for each pixel in an input RGBD image, we cast a ray from the pixel center towards the camera pinhole and use the depth map to determine the point location. 
For each point, we gather the following information (see \cref{fig: architecture details}a for illustration):
\vspace{-0.5em}
\begin{itemize}[leftmargin=*]
    \setlength\itemsep{-.1em}
	\item \textit{xyz}, or $p \in \R^3$, the location of the point;
    \item \textit{rgb}, or $c \, {\in} \, \R^3$, color of the point in input RGBD image;
    \item $\vec{r}_d \, {\in} \, \bbS^2$, the cast camera ray direction (normalized);
    \item $u \in \R^3$, a vector from the point corresponding to the current pixel to the point corresponding to the next pixel in the x direction on the input RGBD image (see \cref{fig: architecture details}a);
    \item $v \in \R^3$, same as $u$ but in the y direction;
    \item $z \in \bbS^2$, the optical axis direction of the input camera;
    \item $\ell_c \in \{0, 1\}$, a binary indicator, which is set to 1 when $c$ contains valid information or $0$ when we set $c$ to $(0.5, 0.5, 0.5)$;
    \item and $\ell_a \in \{0, 1\}$, a binary indicator, which is set to 1 when $\vec{r}_d, u, v, v, z$ contain valid information or $0$ when we set all of them to zeors.
\end{itemize}
\vspace{-0.5em}
All of the features are point-wise and can be easily extracted from the camera pose. 
To support point clouds that contain only $xyz$ information and without these information, we randomly drop $rgb$ and other features independently 50 \% of the time (and set $\ell_c$ and $\ell_a$ accordingly).
During inference, in all experiments shown in the paper, we do not use any of the features, except $xyz$ and $rgb$.

To diversify the sampling rates of the point cloud, we randomly set the resolution (ranging from $30 \times 30$ to $300 \times 300$) and position (1 to 3 units from the origin) for each input camera. 
The target camera has a $50 \times 50$ resolution, \ie, 2500 query rays per iteration. 
To increase the diversity of the query rays, we point the target camera towards a random point in a centered box of a width equal to 1 unit and position the camera at a random location (between 0.5 to 3 units to the origin). 
To help learning the blending weights of color, at every iteration we select a random image patch with a size from $20\times 20$ to $200 \times 200$ in the ImageNet dataset as the texture map for the mesh.
We also select a random $k \in [12, 200]$ at every iteration.  

To optimize the loss function in \cref{eq: training loss}, We use ADAM \cite{kingma2014adam} with $\beta_1 = 0.9$, $\beta_2 = 0.98$, and a learning rate schedule used by \citet{vaswani2017attention} with a warm-up period of 4,000 iterations.
Within the warm-up iterations, the learning rate increases rapidly to $2e^{-6}$, and it gradually decreases afterwards.
We train the model for 350,000 iterations, and it takes 10 days on 8 A100 GPUs.

\section{Inverse rendering: details}
\label{sec: inverse rendering details}

As we have discussed, the pointersect model $f$ allows gradient computation of color and surface normal with respect to the point cloud's $xyz$ and $rgb$.
In this section, we provide details of our inverse rendering experiment in \Cref{sec: exp inverse}.
Our goal is to demonstrate the use of pointersect in an inverse rendering application, so we assume a simple scenario where we have ground-truth RGB images, camera poses, and foreground segmentation masks.
Our purpose is to demonstrate the potential, not to compare with the state-of-the-art inverse rendering methods.

Given $N$ input RGB images, $\mI = \{ I_1, \dots, I_N | I_i \in \R^{h \times w  \times 3} \}$, their corresponding foreground masks,  $\mY = \{ Y_1, \dots, Y_N | Y_i \in [0,1]^{h \times w} \}$, camera extrinsic and intrinsic matrices, and a noisy point cloud, $\cP = \{ (p_1, c_1), \dots, (p_n, c_n) \}$, our goal is to optimize the position $\p = \{p_1, \dots, p_n \}$ and color $\c = \{c_1, \dots, c_n \}$ of the points such that when we render $\cP$ with pointersect from the input camera views, the output images match the input ones.

Let $\r^j_k$ be a camera ray from the $j$-th input image. 
We use bilinear interpolation to calculate the corresponding color and the foreground mask values, $\hat{c}(\r_k^j)$ and $\hat{y}(\r_k^j)$, respectively.
We optimize the following loss function while fixing the network parameter of the pointersect model (\ie, simply use it as part of the rendering forward function): 
\begin{align}
	& \min_{\cP} \sum_{j=1}^{N} \sum_{\r_k^j} \left\| \hat{c}(\r_k^j) -  c(\r_k^j, \cP)   \right\|^2_2  \label{eq: inv loss} \\
	& - \hat{y}(\r_k^j) \log  h(\r_k^j, \cP)  \label{eq: inv hit 1} \\
	& - (1 - \hat{y}(\r_k^j)) \log  (1 - h(\r_k^j, \cP))  \label{eq: inv hit 2} \\
	& + \left(n(\r_k^j, \cP) \cross n(\r_{k+1_u}^j, \cP) \right)^2  \label{eq: inv normal 1} \\
	& + \left(n(\r_k^j, \cP) \cross n(\r_{k+1_v}^j, \cP) \right)^2,  \label{eq: inv normal 2}
\end{align}
where $c(\r_k^j, \cP)$, $n(\r_k^j, \cP)$, and $h(\r_k^j, \cP)$ are pointersect's estimates of color (using blending weights), surface normal, and hit, respectively (see \Cref{sec: pointersect} in the main paper).  
In the optimization program, we minimize the $\ell_2$ loss of color (\equref{eq: inv loss}), the negative log-likelihood of foreground hit estimation (\equref{eq: inv hit 1} and \equref{eq: inv hit 2}), and the smoothness of the estimated normal map between neighboring pixels on the images (\equref{eq: inv normal 1} and \equref{eq: inv normal 2}).
The loss terms related to normal smoothness is optional---we add it for demonstration. 
Note that we do not couple the input $rgb$ colors with the ground-truth surface normal to provide additional supervision to the geometry---the color is computed simply by interpolating the $rgb$ values of input images.

We create an example problem by capturing 100 RGBD images from the Stanford Bunny mesh \cite{Texturemontage05} (whose size is scaled to have 2 units in length), adding Gaussian noise with standard deviation equal to $0.2$ to the depth channel, and using the RGBD images to create the noisy point cloud and RGB images as our input images. 
We solve the optimization problem using stochastic gradient descent with a learning rate of 0.01 with 10000 iterations. 
Each iteration we select a point in $\cP$, project it onto all input images, and cast rays from the local $10 \times 10$ patch of the projected pixel location.
We cast roughly a total of 10000 rays every iteration.
At the end of every iteration, we project the RGB values of the point cloud to $[0, 1]$.
Every 150 iterations, we use the ground-truth foreground segmentation maps to perform silhouette carving on $\cP$, which allows us to remove points that are apparently invalid.
We insert new points by casting rays from random input pixels using pointersect's estimates of point position and color.
This operation is fast, easy to implement, and significantly speeds up the optimization.
The entire optimization (10000 iterations) takes 1 hour on one A100 GPU.

\begin{figure}[t]
	\centering
	\includegraphics[width=\linewidth]{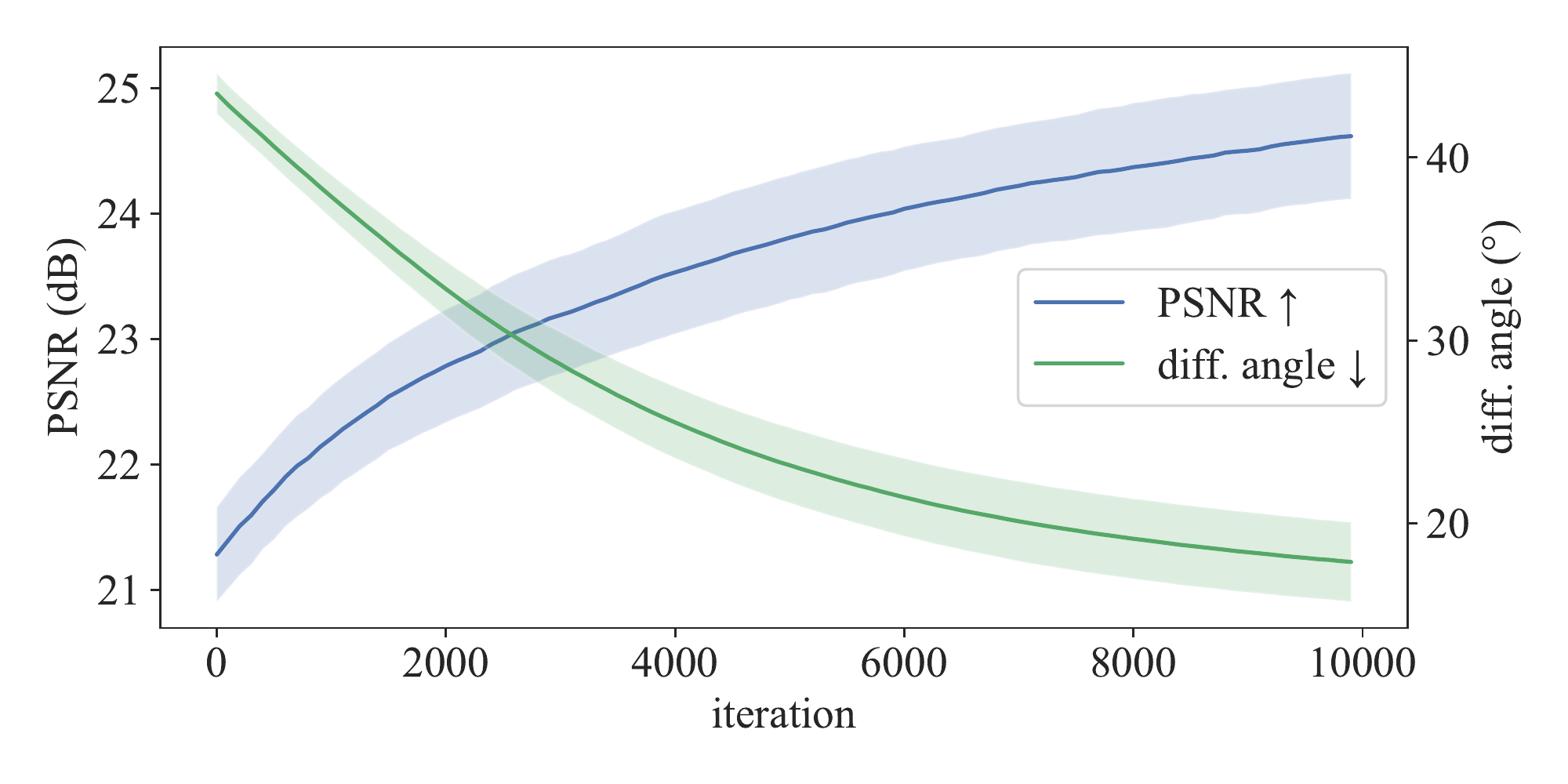}	
	\vspace{-8mm}
	\caption{PSNR and the angle between the ground-truth and estimated color and surface normal, respectively, during the inverse rendering optimization. The shade represents the standard deviation of the values over the 100 input views.} 
	\label{fig: inv curve}
\end{figure}

\begin{table}[t]
	\caption{Optimize a noisy point cloud with pointersect.}
	\label{table: inv}
	\vspace{-3mm}
	\centering
	\begin{adjustbox}{max width=\linewidth}
		\begin{tabular}{lcccc}
			\toprule
			& %
			\multicolumn{2}{c}{100 input views} & 
			\multicolumn{2}{c}{144 novel views} 
			\\
			& 
			before opt.& 
			after opt. & 
			before opt. & 
			after opt. 
			\\
			\midrule
			PSNR (dB) $\uparrow$ & 
			$10.1 \pm 0.5$&   %
			$24.6 \pm 0.5$ &  %
			$13.9 \pm 0.8$ &   %
			$25.7 \pm 1.2$   %
			\\  
			normal (angle (\degree)) $\, \downarrow$ & 
			$54.0 \pm 0.9$&   %
			$17.8 \pm 2.1$ &  %
			$55.3 \pm 0.4$ &   %
			$18.3 \pm 2.7$    %
			\\  
			depth (rmse) $\, \downarrow$ & 
			$0.46 \pm 0.08$&   %
			$0.09 \pm 0.06$ &  %
			$0.33 \pm 0.08$ &   %
			$0.10 \pm 0.07$   %
			\\  
			\bottomrule
		\end{tabular}
	\end{adjustbox}
	 \vspace{-2mm}
\end{table}

\begin{figure*}[t]
	\centering
	\includegraphics[width=\linewidth]{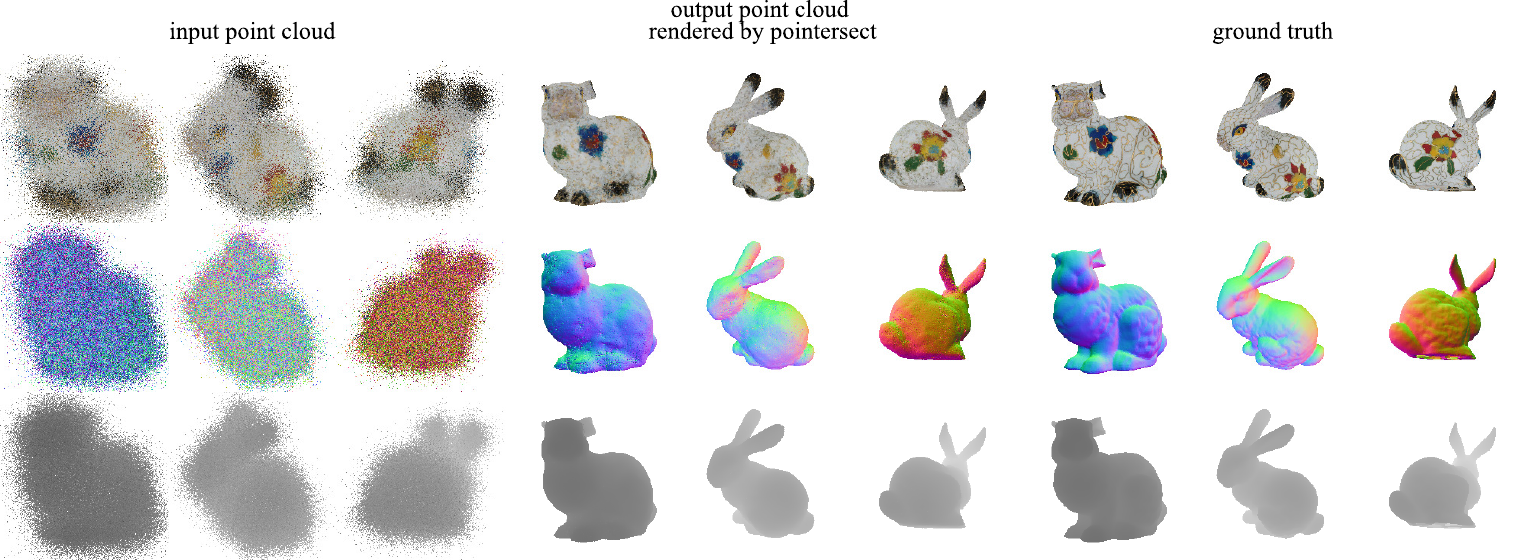}	
	\vspace{-6mm}
	\caption{We back-propagate gradient through the pointersect model to optimize a noisy point cloud, comparing output images with the given clean RGB images and foreground binary mask.} 
	\label{fig: inv detail}
\end{figure*}

\autoref{fig: inv detail} shows the results, \cref{fig: inv curve} shows the optimization progression, and \autoref{table: inv} shows the statistics on 100 input views and 144 novel views.  
As can be seen, with the capability to back-propogate gradient through pointersect, we are able to optimize the point cloud to reduce the rgb, depth, and normal errors.

\section{Deviation from ground-truth surface}
\label{sec: chamfer}
We measure the error ``along'' the ray as we are interested in the accuracy from a ray intersection stand point, which is the main focus of the paper.
To further provide insight into the accuracy of the surfaces that would be reconstructed, \autoref{table: chamfer} shows the Chamfer distance using the same setup as \autoref{table: basic}.

\begin{table}[h]
	\caption{ Chamfer distance ($\times 10^{-3}$).  Note that Chamfer distance measures square distances, whereas the RMSE (used by the rest of the paper) measure distance. 
	}
	\label{table: chamfer}
		\vspace{-1mm}
		\centering
		\begin{adjustbox}{max width=\linewidth}
			\begin{tabular}{lcccc}
				\toprule
				dataset & Visibility splatting &  Poisson recon. & Neural Points  & Ours 
				\\
				\midrule 
				ShapeNet & 
				$2.32 \pm 9.55$ & 
				$0.51 \pm 1.98$ & 
				$0.30 \pm 0.43$ & 
				$\mathbf{0.29 \pm 0.48}$ \\
				Sketchfab & 
				$13.33 \pm 25.53$ & 
				$1.58 \pm 4.30$ & 
				$0.98 \pm 2.47$ & 
				$\mathbf{0.87 \pm 2.19}$ \\  
				\bottomrule
			\end{tabular}
		\end{adjustbox}
		\vspace{-4mm}
\end{table}

\section{Additional results}
\label{sec: additional results}

\begin{figure*}[t]
	\centering
	\begin{subfigure}[t]{\linewidth}
		\centering
		\includegraphics[width=\linewidth]{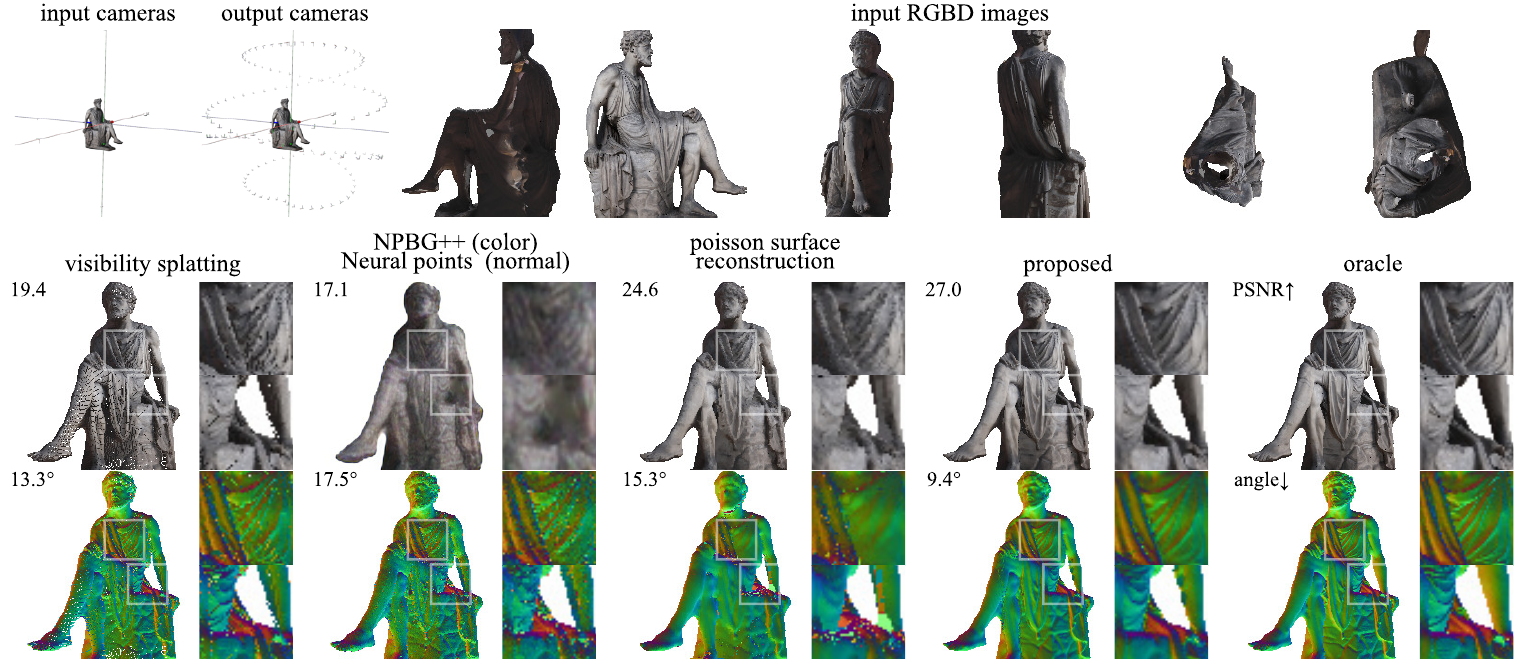}
	\end{subfigure}
	\\[1em]
	\begin{subfigure}[t]{\linewidth}
		\centering
		\includegraphics[width=\linewidth]{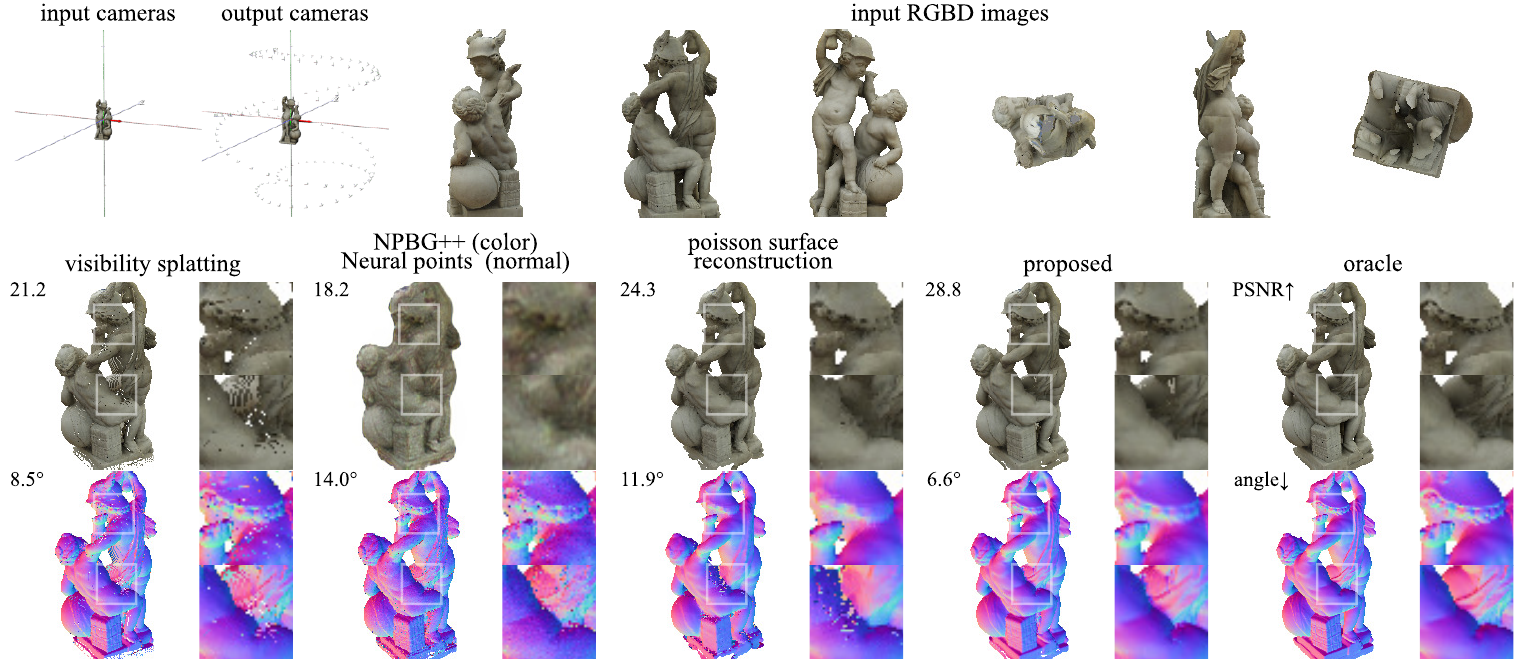}
	\end{subfigure}
	\caption{Additional example results of pointersect and baselines on Sketchfab dataset.  Mesh credit: \ccLogo~\citet{seatedjew}.  \ccLogo~\citet{cupid}. 
 }
	\label{fig: additional basic}
\end{figure*}

\Cref{fig: additional basic} extends \Cref{fig: basic} in the paper and shows two results from the Sketchfab test dataset. 
The supplementary offline website showcases the result videos of \Cref{sec: exp basics}, \Cref{sec: exp hypersim}, \Cref{sec: exp inverse}, and \Cref{sec: exp real}. 
In the following, we provide two ablation studies on the number of views, the choice of $k$, and the sampling rate (\ie, density) of the point clouds.

\subsection{Number of input views and the choice of $k$}
\label{sec: num views and k}
In \Cref{sec: exp basics}, we constructed the input point clouds by capturing 6 $200 \times 200$ RGBD images, each from front, back, left, right, top, and bottom, of the object of interest.
The input point cloud, constructed in this manner, may fail to contain occluded part of the object.
Here, we repeat the experiment with an increased number of input RGBD images, 30 and 60.
Specifically, we randomly sample 30 (or 60) cameras uniformly within a sphere shell with inner radius of 3 and outer radius of 4.
All cameras have a field of view of 30 degrees, a resolution of $200 \times 200$, and point to the center of the object of interest.
All settings are the same as those in \Cref{sec: exp basics}---$k=40$, $\delta=0.1$, and we provide ground-truth vertex normal to Poisson reconstruction and visibility splatting.
The depth and normal errors are computed only on non-hole pixels.
Additionally, since increasing the number of input views increases the density of the point cloud, choosing the same number of nearest points, \ie, using the same $k$, effectively reduces the size of the neighborhood the pointersect model can attend to.
Thus, we also include a pointersect result where we increase $k$ to 200, which retains the expected neighborhood size for 30 input views and halves it for 60 input views.

The results are shown in \cref{table: basic in30}.
As can be seen, pointersect, regardless of the choice of $k$, achieves the best results in color estimations.
Increasing $k$ (thus allows pointersect to attend to a larger neighborhood) also makes pointersect outperform all baselines in terms of the estimation accuracy of depth and surface normal.

\begin{table*}[t]
	\caption{Test results on three datasets. Inputs are 30 or 60 RGBD images captured at random locations within sphere shell, pointing toward the center. All test meshes are unseen during training.  NGP does not use depth information and is trained for 1000 epochs (about 10 minutes). It is included as a reference baseline.}
	\label{table: basic in30}
	\centering
	\begin{adjustbox}{max width=\linewidth}
		\begin{tabular}{llllllll}
			\toprule
			\multirow{2.5}{*}{Method} & 
			\multirow{2.5}{*}{Metrics} & 
			\multicolumn{2}{c}{tex-models} & 
			\multicolumn{2}{c}{ShapeNet} &
			\multicolumn{2}{c}{Sketchfab}
			\\
			\cmidrule(lr){3-4} \cmidrule(lr){5-6} \cmidrule(lr){7-8}  
			&
			&
			\makecell{30 input views} & 
			\makecell{60 input views} & 
			\makecell{30 input views} &
			\makecell{60 input views} &
			\makecell{30 input views} &
			\makecell{60 input views} 
			\\
			\midrule

			\multirow{6}{4em}{Visibility splatting}  %
			& depth (RMSE) $\, \downarrow$ & $0.06 \pm 0.07$ & $0.04 \pm 0.03$ & $0.03 \pm 0.03$ & $0.03 \pm 0.03$ & $0.05 \pm 0.03$ & $0.04 \pm 0.02$ \\
			& normal (angle (\degree)) $\, \downarrow$ & $5.59 \pm 2.40$ & $5.72 \pm 2.64$ & $\mathbf{6.85 \pm 3.67}$ & $\mathbf{7.01 \pm 3.82}$ & $7.84 \pm 3.06$ & $8.13 \pm 3.31$ \\
			& hit (accuracy (\%)) $\, \uparrow$ & $99.1 \pm 0.3$ & $99.0 \pm 0.3$ & $99.1 \pm 0.5$ & $99.1 \pm 0.6$ & $99.3 \pm 0.2$ & $99.2 \pm 0.2$ \\
			& color (PSNR (dB))$\, \uparrow$ & $22.0 \pm 1.8$ & $21.9 \pm 2.1$ & $24.0 \pm 3.0$ & $23.8 \pm 3.1$ & $23.3 \pm 1.3$ & $23.2 \pm 1.5$ \\
			& color (SSIM)$\, \uparrow$ & $0.9 \pm 0.0$ & $0.9 \pm 0.0$ & $0.9 \pm 0.1$ & $0.9 \pm 0.1$ & $0.9 \pm 0.1$ & $0.9 \pm 0.1$ \\
			& color (LPIPS)$\, \downarrow$ & $0.07 \pm 0.05$ & $0.05 \pm 0.02$ & $0.05 \pm 0.03$ & $0.04 \pm 0.03$ & $0.08 \pm 0.06$ & $0.07 \pm 0.03$ \\
			\graymidrule
			\multirow{6}{4em}{Poisson surface recon.}  %
			& depth (RMSE) $\, \downarrow$ & $0.02 \pm 0.04$ & $0.02 \pm 0.04$ & $0.03 \pm 0.07$ & $0.03 \pm 0.06$ & $0.06 \pm 0.09$ & $0.06 \pm 0.09$ \\
			& normal (angle (\degree)) $\, \downarrow$ & $5.76 \pm 2.65$ & $5.46 \pm 2.53$ & $11.14 \pm 6.04$ & $10.49 \pm 6.89$ & $10.59 \pm 6.51$ & $10.14 \pm 6.18$ \\
			& hit (accuracy (\%)) $\, \uparrow$ & $\mathbf{99.9 \pm 0.1}$ & $\mathbf{99.9 \pm 0.1}$ & $98.1 \pm 7.3$ & $99.3 \pm 1.3$ & $99.6 \pm 0.4$ & $99.6 \pm 0.6$ \\
			& color (PSNR (dB))$\, \uparrow$ & $27.5 \pm 3.1$ & $27.7 \pm 3.1$ & - & - & $26.5 \pm 3.6$ & $26.5 \pm 3.9$ \\
			& color (SSIM)$\, \uparrow$ & $0.9 \pm 0.0$ & $0.9 \pm 0.0$ & $0.9 \pm 0.1$ & $0.9 \pm 0.0$ & $0.9 \pm 0.0$ & $0.9 \pm 0.0$ \\
			& color (LPIPS)$\, \downarrow$ & $0.05 \pm 0.03$ & $0.05 \pm 0.03$ & $0.05 \pm 0.06$ & $0.05 \pm 0.04$ & $0.08 \pm 0.04$ & $0.08 \pm 0.04$ \\
			\graymidrule
			\multirow{6}{4em}{Neural points \cite{feng2022np}}  %
			& depth (RMSE) $\, \downarrow$ & $0.04 \pm 0.02$ & $0.04 \pm 0.02$ & $0.03 \pm 0.03$ & $0.03 \pm 0.03$ & $0.04 \pm 0.02$ & $0.04 \pm 0.02$ \\
			& normal (angle (\degree)) $\, \downarrow$ & $12.59 \pm 2.73$ & $13.00 \pm 2.58$ & $16.93 \pm 4.57$ & $16.86 \pm 4.44$ & $15.24 \pm 3.21$ & $15.64 \pm 3.15$ \\
			& hit (accuracy (\%)) $\, \uparrow$ & $99.0 \pm 0.4$ & $99.0 \pm 0.4$ & $99.0 \pm 0.7$ & $99.0 \pm 0.7$ & $99.2 \pm 0.2$ & $99.2 \pm 0.2$ \\
			& color (PSNR (dB))$\, \uparrow$ & not supp. & not supp. & not supp. & not supp. & not supp. & not supp. \\
			& color (SSIM)$\, \uparrow$ & not supp. & not supp. & not supp. & not supp. & not supp. & not supp. \\
			& color (LPIPS)$\, \downarrow$ & not supp. & not supp. & not supp. & not supp. & not supp. & not supp. \\
			\graymidrule
			\multirow{6}{4em}{NPBG++ \cite{rakhimov2022npbg++}}  %
			& depth (RMSE) $\, \downarrow$ & not supp. & not supp. & not supp. & not supp. & not supp. & not supp. \\
			& normal (angle (\degree)) $\, \downarrow$ & not supp. & not supp. & not supp. & not supp. & not supp. & not supp. \\
			& hit (accuracy (\%)) $\, \uparrow$ & not supp. & not supp. & not supp. & not supp. & not supp. & not supp. \\
			& color (PSNR (dB))$\, \uparrow$ & $17.2 \pm 2.4$ & $17.3 \pm 2.4$ & $19.8 \pm 4.1$ & $19.8 \pm 4.1$ & $18.8 \pm 1.7$ & $18.9 \pm 1.7$ \\
			& color (SSIM)$\, \uparrow$ & $0.7 \pm 0.1$ & $0.7 \pm 0.1$ & $0.8 \pm 0.1$ & $0.8 \pm 0.1$ & $0.8 \pm 0.1$ & $0.8 \pm 0.1$ \\
			& color (LPIPS)$\, \downarrow$ & $0.23 \pm 0.05$ & $0.23 \pm 0.05$ & $0.17 \pm 0.08$ & $0.17 \pm 0.08$ & $0.21 \pm 0.07$ & $0.21 \pm 0.06$ \\
			\graymidrule
			\multirow{6}{4em}{NGP \cite{mueller2022instant}}  %
			& depth (RMSE) $\, \downarrow$ & not supp. & not supp. & not supp. & not supp. & not supp. & not supp. \\
			& normal (angle (\degree)) $\, \downarrow$ & not supp. & not supp. & not supp. & not supp. & not supp. & not supp. \\
			& hit (accuracy (\%)) $\, \uparrow$ & not supp. & not supp. & not supp. & not supp. & not supp. & not supp. \\
			& color (PSNR (dB))$\, \uparrow$ & $18.3 \pm 7.7$ & $22.3 \pm 8.2$ &  - & - & $23.7 \pm 6.9$ & $24.8 \pm 7.3$ \\
			& color (SSIM)$\, \uparrow$ & $0.8 \pm 0.1$ & $0.8 \pm 0.2$ & $0.9 \pm 0.1$ & $0.9 \pm 0.1$ & $0.9 \pm 0.1$ & $0.9 \pm 0.1$ \\
			& color (LPIPS)$\, \downarrow$ & $0.22 \pm 0.14$ & $0.16 \pm 0.19$ & $0.14 \pm 0.10$ & $0.17 \pm 0.14$ & $0.14 \pm 0.10$ & $0.12 \pm 0.11$ \\
			\graymidrule
			\multirow{6}{4em}{Proposed ($k \,{=}\,40$)}  %
			& depth (RMSE) $\, \downarrow$ & $0.02 \pm 0.02$ & $0.02 \pm 0.02$ & $0.02 \pm 0.02$ & $0.02 \pm 0.02$ & $0.03 \pm 0.02$ & $0.03 \pm 0.02$ \\
			& normal (angle (\degree)) $\, \downarrow$ & $5.85 \pm 1.82$ & $6.30 \pm 1.75$ & $8.85 \pm 3.61$ & $9.25 \pm 3.29$ & $7.01 \pm 2.11$ & $7.47 \pm 2.02$ \\
			& hit (accuracy (\%)) $\, \uparrow$ & $\mathbf{99.9 \pm 0.0}$ & $\mathbf{99.9 \pm 0.0}$ & $\mathbf{99.8 \pm 0.3}$ & $\mathbf{99.8 \pm 0.2}$ & $\mathbf{99.9 \pm 0.0}$ & $\mathbf{99.9 \pm 0.0}$ \\
			& color (PSNR (dB))$\, \uparrow$ & $\mathbf{32.4 \pm 2.1}$ & $\mathbf{33.3 \pm 1.9}$ & $\mathbf{29.7 \pm 3.8}$ & $\mathbf{30.5 \pm 3.9}$ & $30.3 \pm 2.7$ & $30.9 \pm 2.7$ \\
			& color (SSIM)$\, \uparrow$ & $\mathbf{1.0 \pm 0.0}$ & $\mathbf{1.0 \pm 0.0}$ & $\mathbf{1.0 \pm 0.0}$ & $\mathbf{1.0 \pm 0.0}$ & $\mathbf{1.0 \pm 0.0}$ & $\mathbf{1.0 \pm 0.0}$ \\
			& color (LPIPS)$\, \downarrow$ & $0.02 \pm 0.02$ & $\mathbf{0.01 \pm 0.02}$ & $\mathbf{0.03 \pm 0.03}$ & $\mathbf{0.02 \pm 0.02}$ & $\mathbf{0.04 \pm 0.04}$ & $\mathbf{0.04 \pm 0.04}$ \\
			\graymidrule
			\multirow{6}{4em}{Proposed ($k \,{=} \, 200$)}  %
			& depth (RMSE) $\, \downarrow$ &$\mathbf{0.01 \pm 0.01}$ & $\mathbf{0.01 \pm 0.01}$ & $\mathbf{0.01 \pm 0.01}$ & $\mathbf{0.01 \pm 0.01}$ & $\mathbf{0.01 \pm 0.01}$ & $\mathbf{0.01 \pm 0.01}$ \\
			& normal (angle (\degree)) $\, \downarrow$ & $\mathbf{5.09 \pm 1.83}$ & $\mathbf{4.91 \pm 1.65}$ & $8.08 \pm 3.57$ & $7.59 \pm 3.38$ & $\mathbf{5.88 \pm 1.96}$ & $\mathbf{5.43 \pm 1.69}$ \\
			& hit (accuracy (\%)) $\, \uparrow$ & $\mathbf{99.9 \pm 0.1}$ & $\mathbf{99.9 \pm 0.0}$ & $99.7 \pm 0.3$ & $\mathbf{99.8 \pm 0.3}$ & $\mathbf{99.9 \pm 0.1}$ & $\mathbf{99.9 \pm 0.0}$ \\
			& color (PSNR (dB))$\, \uparrow$ & $\mathbf{32.4 \pm 2.6}$ & $\mathbf{33.3 \pm 2.6}$ & $\mathbf{29.7 \pm 3.6}$ & $30.4 \pm 3.8$ & $\mathbf{30.9 \pm 2.7}$ & $\mathbf{31.6 \pm 2.8}$ \\
			& color (SSIM)$\, \uparrow$ & $\mathbf{1.0 \pm 0.0}$ & $\mathbf{1.0 \pm 0.0}$ & $\mathbf{1.0 \pm 0.0}$ & $\mathbf{1.0 \pm 0.0}$ & $\mathbf{1.0 \pm 0.0}$ & $\mathbf{1.0 \pm 0.0}$ \\
			& color (LPIPS)$\, \downarrow$ & $\mathbf{0.01 \pm 0.01}$ & $\mathbf{0.01 \pm 0.00}$ & $\mathbf{0.03 \pm 0.03}$ & $\mathbf{0.02 \pm 0.02}$ & $\mathbf{0.04 \pm 0.04}$ & $\mathbf{0.04 \pm 0.04}$ \\

			\bottomrule
		\end{tabular}
	\end{adjustbox}
	\vspace{-2mm}
\end{table*}

\subsection{Number of views \vs point density}
\label{sec: num views frontal}
In this experiment, we study the effect of point density when we capture the scene in a frontal-view position.
We take the Stanford Bunny \cite{Texturemontage05} and capture 1 RGBD image from the center frontal view in various resolutions.
Since the field-of-view of remain the same, increasing the resolution increases the density of the points.
We also capture 4 more RGBD images at the 4 corners of the axis-aligned bounding box, also from the frontal view.
The output cameras are on a circle centered at the center viewpoint and looking towards the center of the bunny. 
All settings are the same as those in \Cref{sec: exp basics}, \ie, $k=40$, $\delta=0.1$, and we provide the ground-truth vertex normal to Poisson surface reconstruction and visibility splatting.

\begin{figure*}[t]
	\centering
	\includegraphics[width=\linewidth]{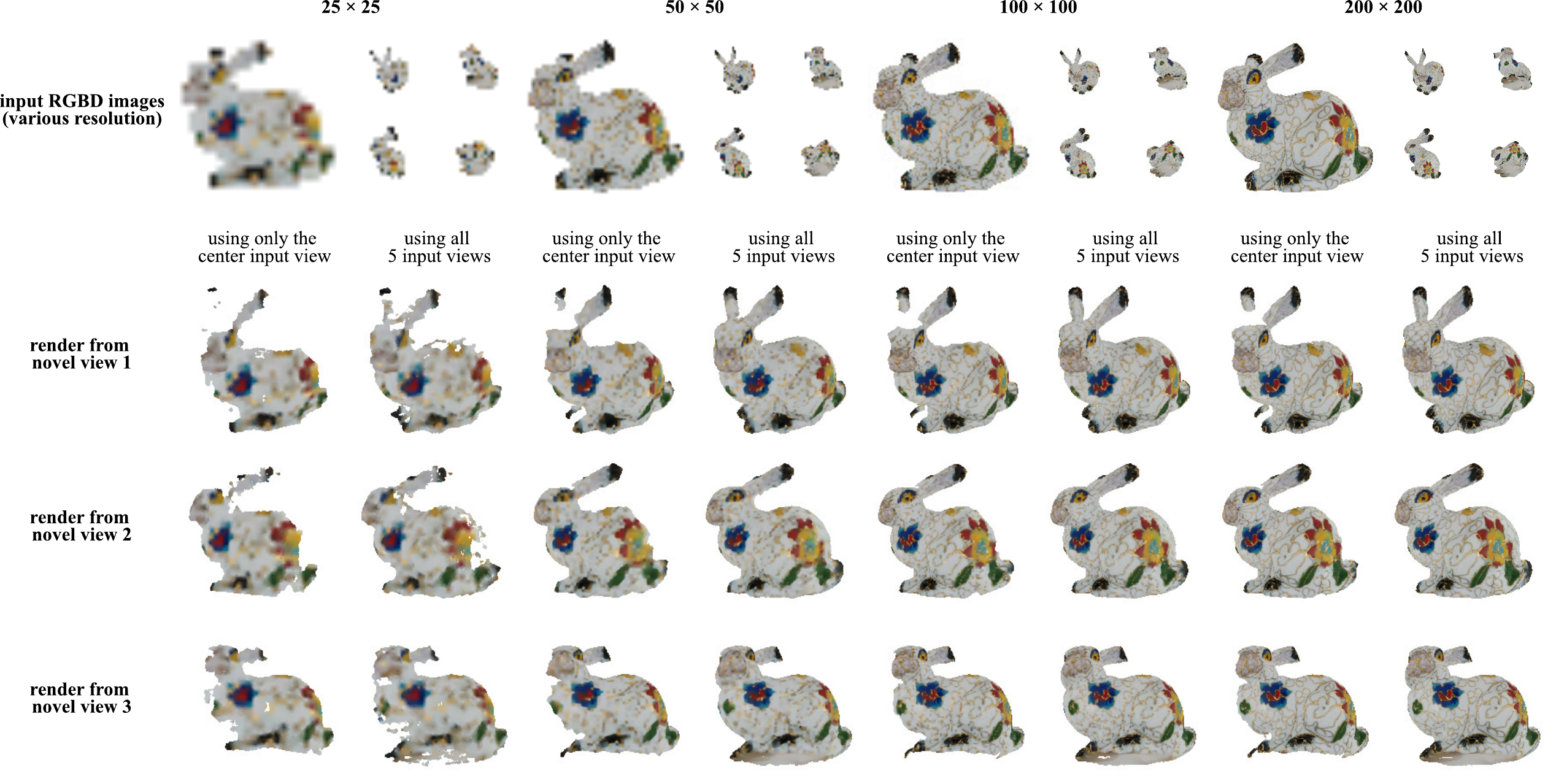}
	\caption{The effect of number of views and the sampling rate of the input point cloud.  We create various kinds of input point clouds of the Stanford Bunny \cite{Texturemontage05} by capturing 1 or 5 input RGBD images of various resolutions. 
	The first row shows the 5 input RGBD images (with the center one enlarged). 
	The input point cloud becomes denser from left to right when the resolution of the input images increases.  With 5 input views, the point cloud covers a wider area than using only the center input view.
	The bottom three rows show the output rendering from pointersect at 3 novel viewpoints.  
	}
	\label{fig: number of views vs resolution}
\end{figure*}

The results are shown in \Cref{table: frontal bunny in1} and \Cref{fig: number of views vs resolution}.
As can be seen, even with a single view point, pointersect is able to render the point cloud from novel view points.
Increasing the point density improves the estimation accuracy of pointersect.
When we use only a single view (and part of the bunny is occluded and missing in the input point cloud), pointersect renders faithfully the point cloud and thus produces with a missing ear. 
The occluded ear appears when we include all 5 input images.

\begin{table*}[t]
	\caption{Test results on the Stanford Bunny mesh with a single input RGBD image captured at the frontal view.  Target views are a frontal circle looking towards the mesh.}
	\label{table: frontal bunny in1}
	\vspace{-2mm}
	\centering
	\begin{adjustbox}{max width=\linewidth}
		\begin{tabular}{llllllllll}
			\toprule
			\makecell{Method} &
			\makecell{Metrics} &
			\makecell{1 view \\ $25 \times 25$}  &  
			\makecell{1 view \\ $50 \times 50$} & 
			\makecell{1 view \\ $100 \times 100$} & 
			\makecell{1 view \\ $200 \times 200$} &
			\makecell{5 views \\ $25 \times 25$}  &  
			\makecell{5 views \\ $50 \times 50$} & 
			\makecell{5 views \\ $100 \times 100$} & 
			\makecell{5 views \\ $200 \times 200$} 
			\\
			\midrule

			\multirow{6}{4em}{Visibility splatting}  %
			& depth (RMSE) $\, \downarrow$ & $0.07 \pm 0.05$ & $0.07 \pm 0.04$ & $0.05 \pm 0.03$ & $0.02 \pm 0.01$ & $0.21 \pm 0.02$ & $0.19 \pm 0.02$ & $0.15 \pm 0.01$ & $0.04 \pm 0.01$ \\
			& normal (angle (\degree)) $\, \downarrow$ & $4.24 \pm 0.44$ & $4.28 \pm 0.34$ & $3.82 \pm 0.26$ & $3.07 \pm 0.10$ & $11.53 \pm 0.87$ & $10.39 \pm 0.82$ & $7.62 \pm 0.56$ & $3.64 \pm 0.22$ \\
			& hit (accuracy (\%)) $\, \uparrow$ & $62.8 \pm 1.2$ & $64.9 \pm 1.2$ & $73.1 \pm 1.1$ & $93.0 \pm 1.0$ & $63.5 \pm 1.2$ & $67.5 \pm 1.2$ & $79.7 \pm 0.9$ & $97.7 \pm 0.1$ \\
			& color (PSNR (dB))$\, \uparrow$ & $11.0 \pm 0.2$ & $11.2 \pm 0.2$ & $12.4 \pm 0.3$ & $18.2 \pm 0.3$ & $11.1 \pm 0.2$ & $11.5 \pm 0.2$ & $13.5 \pm 0.3$ & $21.6 \pm 0.2$ \\
			& color (SSIM)$\, \uparrow$ & $0.6 \pm 0.0$ & $0.6 \pm 0.0$ & $0.6 \pm 0.0$ & $0.8 \pm 0.0$ & $0.6 \pm 0.0$ & $0.6 \pm 0.0$ & $0.6 \pm 0.0$ & $0.8 \pm 0.0$ \\
			& color (LPIPS)$\, \downarrow$ & $0.42 \pm 0.02$ & $0.36 \pm 0.02$ & $0.33 \pm 0.01$ & $0.17 \pm 0.01$ & $0.38 \pm 0.02$ & $0.32 \pm 0.01$ & $0.29 \pm 0.01$ & $0.12 \pm 0.01$ \\
			\graymidrule
			\multirow{6}{4em}{Poisson surface recon.}  %
			& depth (RMSE) $\, \downarrow$ & $0.09 \pm 0.02$ & $0.09 \pm 0.03$ & $0.08 \pm 0.03$ & $0.07 \pm 0.03$ & $0.07 \pm 0.02$ & $0.05 \pm 0.02$ & $0.03 \pm 0.02$ & $0.02 \pm 0.02$ \\
			& normal (angle (\degree)) $\, \downarrow$ & $27.45 \pm 1.45$ & $19.96 \pm 1.28$ & $13.44 \pm 1.67$ & $8.79 \pm 1.53$ & $21.46 \pm 1.07$ & $14.90 \pm 1.00$ & $8.69 \pm 0.69$ & $5.11 \pm 0.44$ \\
			& hit (accuracy (\%)) $\, \uparrow$ & $76.3 \pm 3.3$ & $72.2 \pm 4.8$ & $72.2 \pm 4.8$ & $73.0 \pm 4.9$ & $76.6 \pm 3.2$ & $77.7 \pm 4.5$ & $82.4 \pm 3.9$ & $88.3 \pm 1.2$ \\
			& color (PSNR (dB))$\, \uparrow$ & $12.0 \pm 0.6$ & $10.6 \pm 0.8$ & $11.0 \pm 0.8$ & $11.2 \pm 0.9$ & $11.5 \pm 0.6$ & $12.1 \pm 1.2$ & $13.4 \pm 1.5$ & $15.7 \pm 0.6$ \\
			& color (SSIM)$\, \uparrow$ & $0.4 \pm 0.0$ & $0.4 \pm 0.0$ & $0.5 \pm 0.0$ & $0.6 \pm 0.0$ & $0.4 \pm 0.0$ & $0.5 \pm 0.0$ & $0.6 \pm 0.0$ & $0.8 \pm 0.0$ \\
			& color (LPIPS)$\, \downarrow$ & $0.53 \pm 0.03$ & $0.51 \pm 0.04$ & $0.45 \pm 0.03$ & $0.38 \pm 0.03$ & $0.48 \pm 0.03$ & $0.41 \pm 0.04$ & $0.30 \pm 0.04$ & $0.19 \pm 0.02$ \\
			\graymidrule
			\multirow{6}{4em}{Neural points \cite{feng2022np}}  %
			& depth (RMSE) $\, \downarrow$ & $0.04 \pm 0.01$ & $0.03 \pm 0.01$ & $0.02 \pm 0.00$ & $0.02 \pm 0.00$ & $0.07 \pm 0.01$ & $0.04 \pm 0.01$ & $0.06 \pm 0.01$ & $0.04 \pm 0.01$ \\
			& normal (angle (\degree)) $\, \downarrow$ & $18.01 \pm 0.14$ & $14.94 \pm 0.41$ & $11.83 \pm 0.50$ & $9.73 \pm 0.31$ & $18.37 \pm 0.36$ & $15.04 \pm 0.78$ & $11.91 \pm 0.68$ & $9.87 \pm 0.62$ \\
			& hit (accuracy (\%)) $\, \uparrow$ & $86.9 \pm 1.2$ & $95.1 \pm 1.2$ & $96.3 \pm 1.2$ & $96.8 \pm 1.2$ & $94.4 \pm 0.7$ & $98.1 \pm 0.1$ & $98.3 \pm 0.2$ & $98.7 \pm 0.1$ \\
			& color (PSNR (dB))$\, \uparrow$ & not supp. & not supp. & not supp. & not supp. & not supp. & not supp. & not supp. & not supp. \\
			& color (SSIM)$\, \uparrow$ & not supp. & not supp. & not supp. & not supp. & not supp. & not supp. & not supp. & not supp. \\
			& color (LPIPS)$\, \downarrow$ & not supp. & not supp. & not supp. & not supp. & not supp. & not supp. & not supp. & not supp. \\
			\graymidrule
			\multirow{6}{4em}{Proposed}  %
			& depth (RMSE) $\, \downarrow$ & $0.02 \pm 0.01$ & $0.02 \pm 0.00$ & $0.01 \pm 0.01$ & $0.01 \pm 0.00$ & $0.03 \pm 0.01$ & $0.02 \pm 0.01$ & $0.01 \pm 0.01$ & $0.01 \pm 0.01$ \\
			& normal (angle (\degree)) $\, \downarrow$ & $15.58 \pm 0.45$ & $8.73 \pm 0.31$ & $5.07 \pm 0.10$ & $3.90 \pm 0.10$ & $15.07 \pm 0.69$ & $7.79 \pm 0.52$ & $4.87 \pm 0.26$ & $3.86 \pm 0.16$ \\
			& hit (accuracy (\%)) $\, \uparrow$ & $91.4 \pm 1.0$ & $95.4 \pm 1.2$ & $96.4 \pm 1.2$ & $96.8 \pm 1.2$ & $96.3 \pm 1.0$ & $99.4 \pm 0.1$ & $99.7 \pm 0.1$ & $99.8 \pm 0.1$ \\
			& color (PSNR (dB))$\, \uparrow$ & $16.4 \pm 0.4$ & $19.2 \pm 0.7$ & $20.6 \pm 1.0$ & $21.8 \pm 1.4$ & $18.7 \pm 0.9$ & $22.9 \pm 0.4$ & $25.5 \pm 0.3$ & $28.8 \pm 0.6$ \\
			& color (SSIM)$\, \uparrow$ & $0.6 \pm 0.0$ & $0.7 \pm 0.0$ & $0.8 \pm 0.0$ & $0.9 \pm 0.0$ & $0.7 \pm 0.0$ & $0.8 \pm 0.0$ & $0.9 \pm 0.0$ & $0.9 \pm 0.0$ \\
			& color (LPIPS)$\, \downarrow$ & $0.32 \pm 0.01$ & $0.26 \pm 0.01$ & $0.16 \pm 0.01$ & $0.09 \pm 0.01$ & $0.29 \pm 0.01$ & $0.20 \pm 0.01$ & $0.10 \pm 0.01$ & $0.04 \pm 0.00$ \\

			\bottomrule
		\end{tabular}
	\end{adjustbox}
	\vspace{1mm}
\end{table*}

\subsection{Poisson without ground-truth vertex normal}
\label{sec: gt vertex normal} 
In all experiments in the paper (except the ones on real Lidar point cloud where we do not have ground-truth vertex normal), we provide the ground-truth vertex normal to Poisson surface reconstruction. 
However, in practice, ground-truth vertex normal is difficult to get, and thus we often need to estimate the vertex normal from the point cloud before computing Poisson reconstruction.

\begin{figure*}[t]
	\centering
	\includegraphics[width=0.9\linewidth]{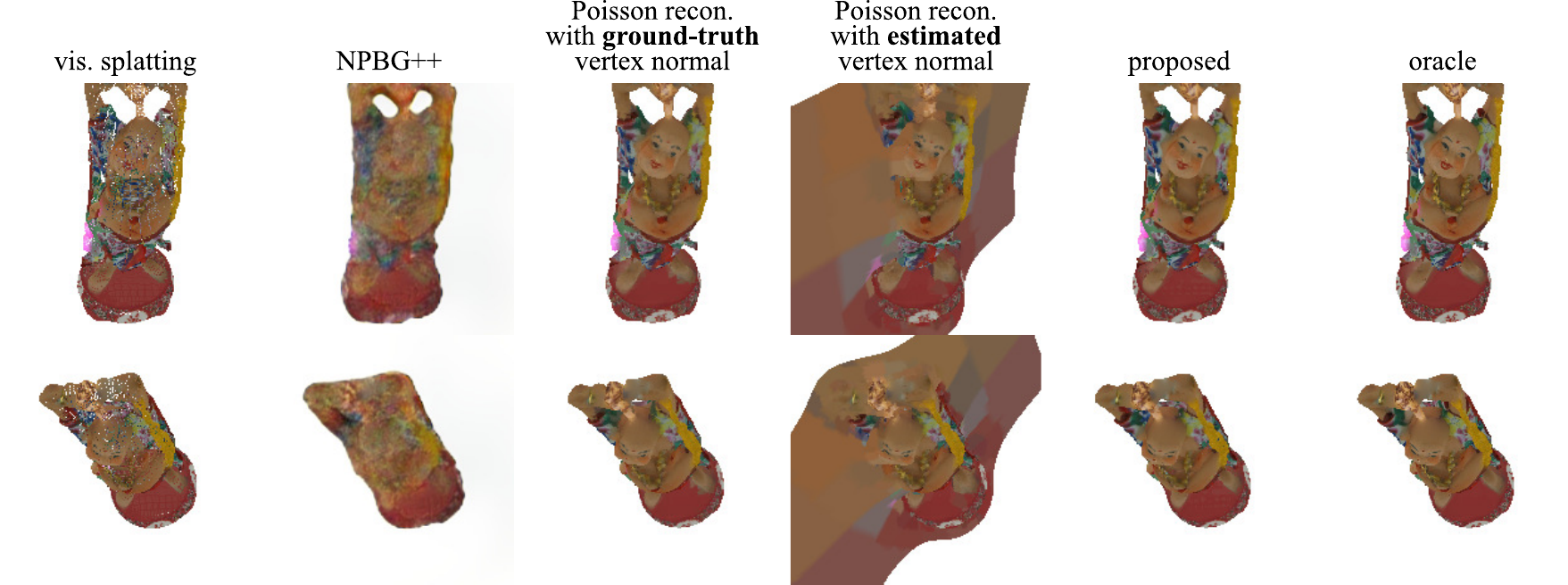}
	\caption{Novel view rendering on Buddha.  The output quality of Poisson surface reconstruction depends highly on the quality of the vertex normal.  Pointersect does not need nor use vertex normal.
	}
	\label{fig: poisson gt or not}
\end{figure*}

In \cref{fig: poisson gt or not} we show an additional result where Poisson reconstruction is given the \textit{estimated} vertex normal.
We estimate the vertex normal directly from the point cloud using  Open3D, which estimates vertex normal by fitting local planes.
As can be seen, the output quality of Possion reconstruction is significantly affected when we use vertex normal that contains a small amount of noise.
Pointersect, on the other hand, does not use vertex normal, so the result is unaffected.

\section{Noisy point cloud from handheld devices}
\label{sec: noisy arkit}

Pointersect is trained on \textit{clean} point clouds that are created from meshes.
Thus, pointersect renders the input point cloud as is---if an input point cloud contains noisy points, the noise will appear in the rendered images and estimated depth and normal maps.
It would be interesting to see what would happen if we directly apply the pointersect model on a noisy point cloud that is captured by a handheld device. 

We take 19 RGBD images from the ARKitScenes dataset~\cite{dehghan2021arkitscenes}, where the depth maps and camera poses are estimated by ARKit.
We perform a simple point-cloud outlier removal utilizing the confidence map output by ARKit.
We then perform a voxel downsampling on the noisy point cloud with a cell width of 0.05. 
We apply various methods to render novel views, including visibility splatting where each point is rendered as 1 pixel, screened Poisson reconstruction, IBRNet~\cite{wang2021ibrnet}, NGP~\cite{mueller2022instant} where we train the model for 200 epochs, and the same pointersect model used in the paper that is trained on clean point clouds.
We use the same setting as those in \Cref{sec: exp real}: $k=100, \delta=0.2$.
Note that Poisson and NGP utilize per-scene optimization; IBRNet does not utilize depth information; pointersect never sees real noisy RGBD images during training. 
It is simply an exploratory experiment to inspire future work.

\begin{figure*}[t]
	\centering
	\begin{subfigure}[t]{\linewidth}
		\centering
		\includegraphics[width=\linewidth]{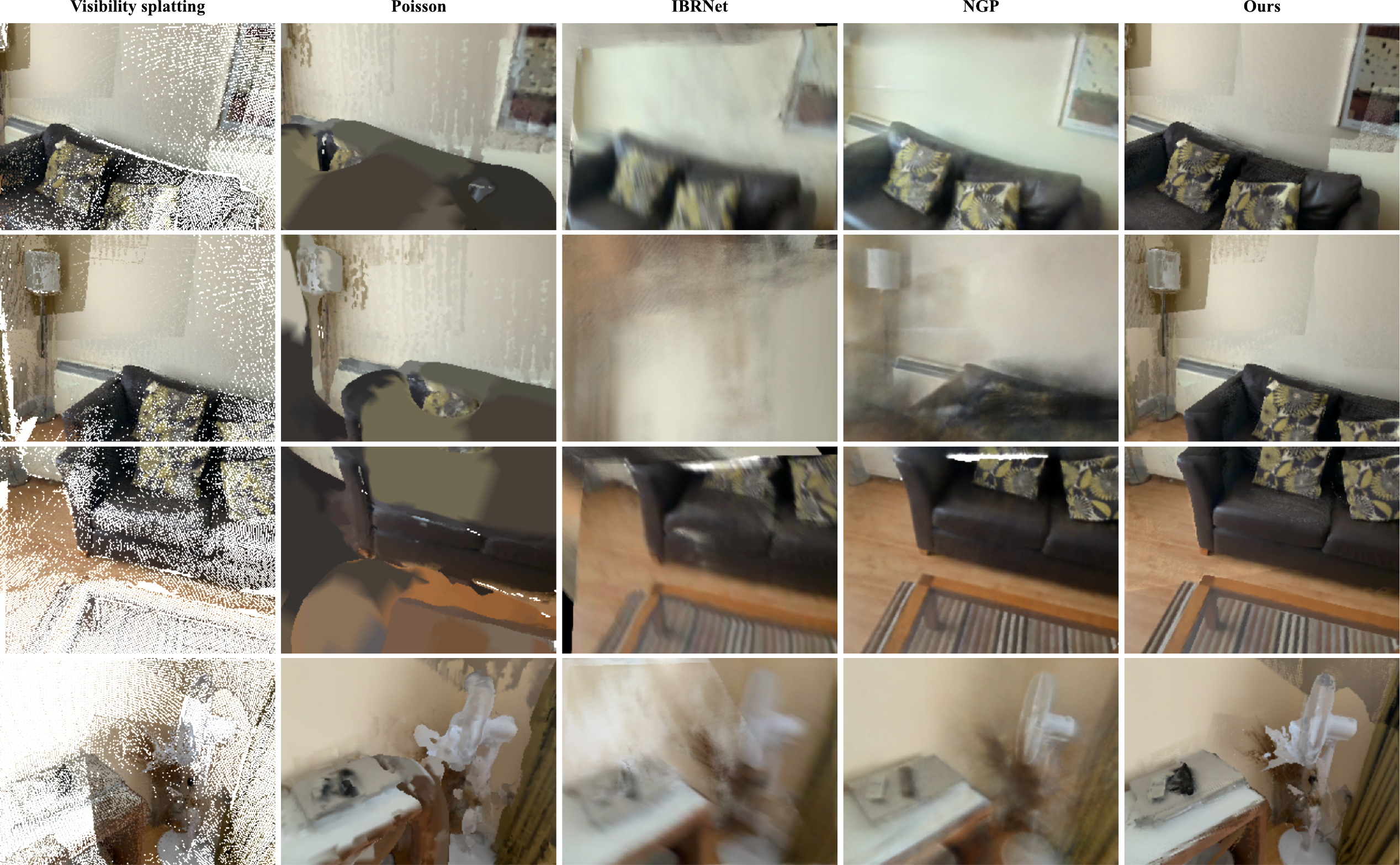}
		\caption{Novel-view rendering on noisy point cloud captured by a handheld device}
	\end{subfigure}
	\\[1em]
	\begin{subfigure}[t]{\linewidth}
		\centering
		\includegraphics[width=\linewidth]{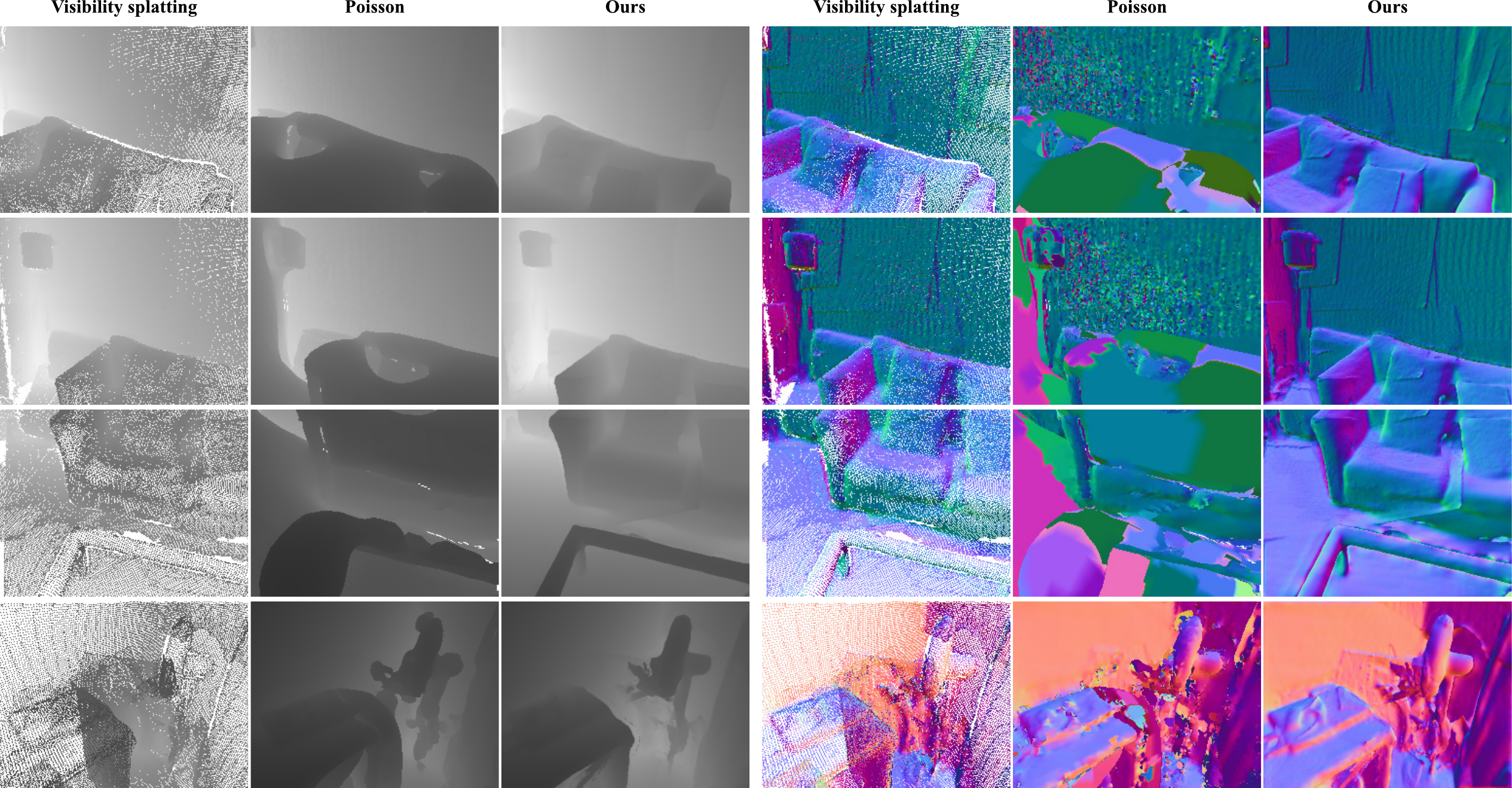}
		\caption{Depth (left) and surface normal (right) estimation at the viewpoints in (a). }
	\end{subfigure}
	\caption{Results on noisy depth maps captured by a handheld device. Inputs are 19 RGBD images from the ARKitScenes dataset~\cite{dehghan2021arkitscenes}, where the depth maps and camera poses are estimated by ARKit and thus contain noise.  Note that the pointersect model is not trained to render noisy point clouds and has never seen real noisy RGBD images during training. The result is provided to motivate future work.
	}
	\label{fig: noisy arkit}
\end{figure*}

\autoref{fig: noisy arkit} shows the results.
As can be seen, while the pointersect model never sees real-world noisy point clouds during training, it is able to render the point cloud with reasonable quality.
Utilizing the depth information, pointersect can directly renders the input images without any training, and the rendered results do not contain floating artifacts.
Since pointersect does not perform any per-scene optimization, the rendered images present the varying exposure and white-balance settings in the input images / point cloud. 
While the surface normal estimation are reasonable, they are affected most severe by the noise in the point cloud and camera poses compared to the estimated depth map and the results when the input point cloud is clean.

\section{Sketchfab dataset}
\label{sec: sketchfab dataset}

\begin{figure*}[t]
	\centering
	\begin{subfigure}[t]{0.79\linewidth}
		\centering
		\includegraphics[height=95mm]{figures/dataset/sketchfab_train.pdf}
		\caption{48 training meshes}
	\end{subfigure}
	\begin{subfigure}[t]{0.19\linewidth}
		\centering
		\includegraphics[height=95mm]{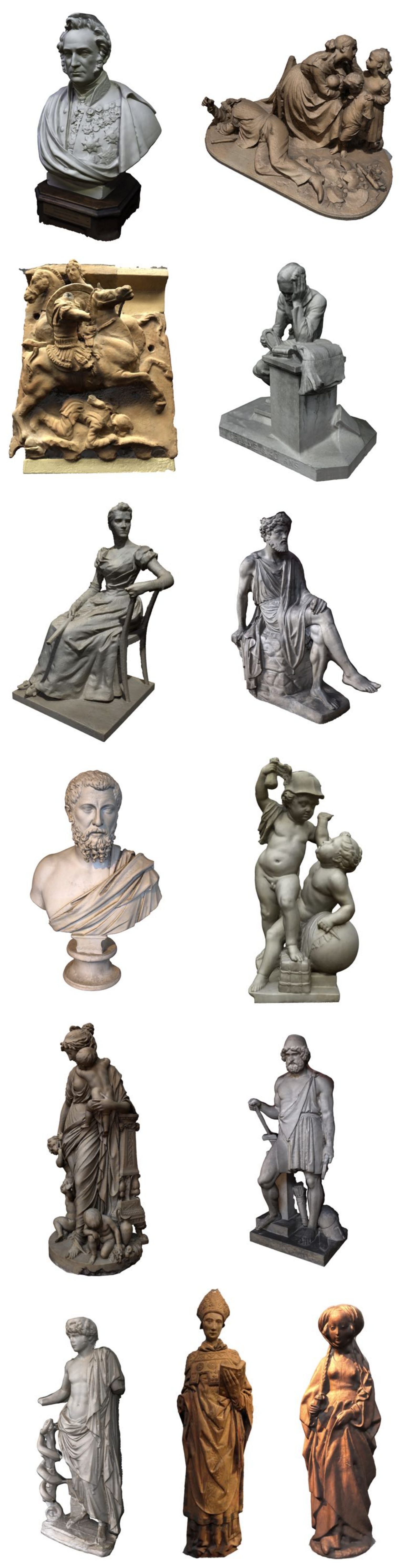}
		\caption{13 test meshes}
	\end{subfigure}
	\caption{(a) The entire training meshes used to train our model. The meshes are from a subset of the Sketchfab dataset \cite{qian2020pugeo}.  (b) is the test meshes.  See \autoref{table: sketchfab composition train} and  \autoref{table: sketchfab composition test} for credits.
	}
	\label{fig: sketchfab composition}
\end{figure*}

We train our model on a subset of the Sketchfab dataset that was used by \citet{qian2020pugeo} to train Neural Points \cite{feng2022np}.
The original dataset contains 90 training meshes and 13 test meshes. 
However, we use only the 48 training meshes that are with variants of the Creative Common license and searchable on \url{sketchfab.com}. 
We use the same 13 test meshes --- they are all with the Creative Common license.
In \autoref{fig: sketchfab composition}, we plot the training and test meshes, and in \autoref{table: sketchfab composition train} and \autoref{table: sketchfab composition test} we list their download links and license information.

\begin{table*}[t]
	\caption{Download links and credits of the entire Sketchfab training set used to train the model.}
	\label{table: sketchfab composition train}
	\vspace{-2mm}
	\centering
	\renewcommand{\arraystretch}{1.25}
	\begin{adjustbox}{max width=\linewidth}
		\begin{tabular}{llp{220mm}}
			\toprule
			\makecell{Name} &
			\makecell{Download link} &
			\makecell{Credit}  
			\\
			\midrule
			Alliance-statue & https://skfb.ly/DHIV & "Warcraft Alliance Statue" (https://skfb.ly/DHIV) by ahtiandr is licensed under Creative Commons Attribution-NonCommercial (http://creativecommons.org/licenses/by-nc/4.0/). \\
			Amphitrite & https://skfb.ly/6n9PF & "Amphitrite - Louvre Museum (Low Definition)" (https://skfb.ly/6n9PF) by Benjamin Bardou is licensed under Creative Commons Attribution (http://creativecommons.org/licenses/by/4.0/). \\
			Ancient-turti & https://skfb.ly/6o9KR & "Archelon - Dragon Sea Turtle" (https://skfb.ly/6o9KR) by inkhero is licensed under Creative Commons Attribution (http://creativecommons.org/licenses/by/4.0/). \\
			Angel5 & https://skfb.ly/6svWA & "Angel Sculpture 3D Scan (Einscan-S)" (https://skfb.ly/6svWA) by 3DWP is licensed under Creative Commons Attribution (http://creativecommons.org/licenses/by/4.0/). \\
			Angel6 & https://skfb.ly/WyIS & "Angels" (https://skfb.ly/WyIS) by rvscanners is licensed under Creative Commons Attribution (http://creativecommons.org/licenses/by/4.0/). \\
			Angel-statue & https://skfb.ly/6qxUQ & "Angel Statue in Fossano" (https://skfb.ly/6qxUQ) by Albyfos is licensed under Creative Commons Attribution (http://creativecommons.org/licenses/by/4.0/). \\
			Angel2 & https://skfb.ly/SNow & "Angel 00" (https://skfb.ly/SNow) by TomaszGap is licensed under CC Attribution-NonCommercial-NoDerivs (http://creativecommons.org/licenses/by-nc-nd/4.0/). \\
			Angel3 & https://skfb.ly/KVRY & "Angel" (https://skfb.ly/KVRY) by Medolino is licensed under Creative Commons Attribution (http://creativecommons.org/licenses/by/4.0/). \\
			Angel4 & https://skfb.ly/SZOv & "Baptismal Angel kneeling" (https://skfb.ly/SZOv) by Geoffrey Marchal is licensed under Creative Commons Attribution-NonCommercial (http://creativecommons.org/licenses/by-nc/4.0/). \\
			Angel-diffuse2 & https://skfb.ly/69usM & "Angel playing harp" (https://skfb.ly/69usM) by OpenScan is licensed under Creative Commons Attribution-NonCommercial (http://creativecommons.org/licenses/by-nc/4.0/). \\
			Armadillo & https://skfb.ly/otzQs & "Stanford Armadillo PBR" (https://skfb.ly/otzQs) by hackmans is licensed under Creative Commons Attribution (http://creativecommons.org/licenses/by/4.0/). \\
			Buddha-sit & https://skfb.ly/6nXwQ & "Buddha" (https://skfb.ly/6nXwQ) by icenvain is licensed under Creative Commons Attribution (http://creativecommons.org/licenses/by/4.0/). \\
			Camera & https://skfb.ly/Lp7U & "Panasonic GH4 Body" (https://skfb.ly/Lp7U) by ScanSource 3D is licensed under Creative Commons Attribution (http://creativecommons.org/licenses/by/4.0/). \\
			Compressor & https://skfb.ly/6oIZS & "Compressor Scan" (https://skfb.ly/6oIZS) by GoMeasure3D is licensed under Creative Commons Attribution (http://creativecommons.org/licenses/by/4.0/). \\
			Dragon-plate & https://skfb.ly/osUnp & "Dragon (decimated sculpt)" (https://skfb.ly/osUnp) by Ashraf Bouhadida is licensed under Creative Commons Attribution (http://creativecommons.org/licenses/by/4.0/). \\
			Dragon- stand & https://skfb.ly/6oTOQ & "Chinese Dragon" (https://skfb.ly/6oTOQ) by icenvain is licensed under Creative Commons Attribution (http://creativecommons.org/licenses/by/4.0/). \\
			Dragon-warrior & https://skfb.ly/otvrO & "Dragon-stl" (https://skfb.ly/otvrO) by Thunk3D 3D Scanner is licensed under Creative Commons Attribution (http://creativecommons.org/licenses/by/4.0/). \\
			Dragon-wing & https://skfb.ly/M9KW & "Wooden Dragon" (https://skfb.ly/M9KW) by jschmidtcreaform is licensed under Creative Commons Attribution (http://creativecommons.org/licenses/by/4.0/). \\
			Dragon2 & https://skfb.ly/BZsM & "XYZ RGB Dragon" (https://skfb.ly/BZsM) by 3D graphics 101 is licensed under Creative Commons Attribution-NonCommercial (http://creativecommons.org/licenses/by-nc/4.0/). \\
			Fox-skull & https://skfb.ly/6UoqE & "Grey Fox skull" (https://skfb.ly/6UoqE) by RISD Nature Lab is licensed under Creative Commons Attribution (http://creativecommons.org/licenses/by/4.0/). \\
			Ganesha & https://skfb.ly/66opT & "The elephant god Ganesha" (https://skfb.ly/66opT) by Geoffrey Marchal is licensed under Creative Commons Attribution-NonCommercial (http://creativecommons.org/licenses/by-nc/4.0/). \\
			Ganesha-plane & https://skfb.ly/ovrr6 & "Ganesha" (https://skfb.ly/ovrr6) by Paulotronics is licensed under Creative Commons Attribution (http://creativecommons.org/licenses/by/4.0/). \\
			Gargo & https://skfb.ly/6S8qQ & "Gargo" (https://skfb.ly/6S8qQ) by rudyprieto is licensed under Creative Commons Attribution (http://creativecommons.org/licenses/by/4.0/). \\
			Grid-dog & https://skfb.ly/DTuH & "Girl With Dog" (https://skfb.ly/DTuH) by pencas is licensed under Creative Commons Attribution (http://creativecommons.org/licenses/by/4.0/). \\
			Golden-elephant & https://skfb.ly/GZSL & "Golden Ephant" (https://skfb.ly/GZSL) by dievitacola is licensed under Creative Commons Attribution-NonCommercial (http://creativecommons.org/licenses/by-nc/4.0/). \\
			Guanyiny & https://skfb.ly/ZuHU & "Guanyin" (https://skfb.ly/ZuHU) by Geoffrey Marchal is licensed under Creative Commons Attribution-NonCommercial (http://creativecommons.org/licenses/by-nc/4.0/). \\
			Happy-vrip & https://skfb.ly/BYQD & "Happy Buddha (Stanford)" (https://skfb.ly/BYQD) by 3D graphics 101 is licensed under Creative Commons Attribution-NonCommercial (http://creativecommons.org/licenses/by-nc/4.0/). \\
			Lion-ball & https://skfb.ly/6GDqU & "King of Narnia: ASLAN" (https://skfb.ly/6GDqU) by Anahit Takiryan is licensed under Creative Commons Attribution (http://creativecommons.org/licenses/by/4.0/). \\
			Man-face & https://skfb.ly/GrGu & "Bust of a Roman" (https://skfb.ly/GrGu) by Geoffrey Marchal is licensed under Creative Commons Attribution-NonCommercial (http://creativecommons.org/licenses/by-nc/4.0/). \\
			Man-statue & https://skfb.ly/T9oz & "Man On Bench Statue PHOTOGRAMMETRY" (https://skfb.ly/T9oz) by MrDavids1 is licensed under Creative Commons Attribution (http://creativecommons.org/licenses/by/4.0/). \\
			Maria & https://skfb.ly/6tzH7 & "Maria Fjodorovna Barjatinskaja" (https://skfb.ly/6tzH7) by Geoffrey Marchal is licensed under Creative Commons Attribution (http://creativecommons.org/licenses/by/4.0/). \\
			Mesh-little-angle & https://skfb.ly/6s8yO & "Little Angle" (https://skfb.ly/6s8yO) by MicMac is licensed under Creative Commons Attribution (http://creativecommons.org/licenses/by/4.0/). \\
			Modello-buddha & https://skfb.ly/6qnIn & "Wooden Buddha statuette" (https://skfb.ly/6qnIn) by andrea.notarstefano is licensed under Creative Commons Attribution (http://creativecommons.org/licenses/by/4.0/). \\
			Mozart & https://skfb.ly/FZyK & "The Infant Mozart" (https://skfb.ly/FZyK) by Geoffrey Marchal is licensed under Creative Commons Attribution (http://creativecommons.org/licenses/by/4.0/). \\
			Roman- sphinx & https://skfb.ly/Y9sx & "Roman Sphinx" (https://skfb.ly/Y9sx) by tony-eight is licensed under Creative Commons Attribution-NonCommercial (http://creativecommons.org/licenses/by-nc/4.0/). \\
			Snake & https://skfb.ly/BrOL & "Cobra Statue" (https://skfb.ly/BrOL) by Jonathan Williamson is licensed under Creative Commons Attribution (http://creativecommons.org/licenses/by/4.0/). \\
			Statue-air-force & https://skfb.ly/LZBS & "a sculpture in Air Force Museum of Vietnam" (https://skfb.ly/LZBS) by HoangHiepVu is licensed under Creative Commons Attribution (http://creativecommons.org/licenses/by/4.0/). \\
			Statue-child-fish & https://skfb.ly/6pJ9s & "Château de Chamarande - France" (https://skfb.ly/6pJ9s) by Sakado is licensed under Creative Commons Attribution (http://creativecommons.org/licenses/by/4.0/). \\
			Statue-death & https://skfb.ly/ovIRF & "The death and the mother" (https://skfb.ly/ovIRF) by Geoffrey Marchal is licensed under Creative Commons Attribution-NonCommercial (http://creativecommons.org/licenses/by-nc/4.0/). \\
			Statue-madona & https://skfb.ly/Lv9v & "Madona Sculpture" (https://skfb.ly/Lv9v) by jan.zachar is licensed under Creative Commons Attribution (http://creativecommons.org/licenses/by/4.0/). \\
			Statue-mother & https://skfb.ly/6pUOp & "Pieta" (https://skfb.ly/6pUOp) by MSU Broad Art Museum is licensed under Creative Commons Attribution (http://creativecommons.org/licenses/by/4.0/). \\
			Statue-napoleon & https://skfb.ly/6xHwD & "Equestrian statue of Napoleon" (https://skfb.ly/6xHwD) by Loïc Norgeot is licensed under Creative Commons Attribution (http://creativecommons.org/licenses/by/4.0/). \\
			Statue- neptune-horse & https://skfb.ly/6npKK & "Neptune - Louvre Museum" (https://skfb.ly/6npKK) by Benjamin Bardou is licensed under Creative Commons Attribution (http://creativecommons.org/licenses/by/4.0/). \\
			Statue-oxen & https://skfb.ly/R9Ps & "Ox Statue (Kek Lok Si Buddhist Temple, Penang)" (https://skfb.ly/R9Ps) by nate\_siddle is licensed under Creative Commons Attribution (http://creativecommons.org/licenses/by/4.0/). \\
			Two-wrestiers-in-combat & https://skfb.ly/RTzv & "Two wrestlers in combat (repost)" (https://skfb.ly/RTzv) by Geoffrey Marchal is licensed under Creative Commons Attribution (http://creativecommons.org/licenses/by/4.0/). \\
			Vase-empire & https://skfb.ly/6uDFR & "Empire vase" (https://skfb.ly/6uDFR) by Geoffrey Marchal is licensed under Creative Commons Attribution-NonCommercial (http://creativecommons.org/licenses/by-nc/4.0/). \\
			Vishnu & https://skfb.ly/6nCoN & "The God Vishnu" (https://skfb.ly/6nCoN) by Geoffrey Marchal is licensed under Creative Commons Attribution (http://creativecommons.org/licenses/by/4.0/). \\
			\bottomrule
		\end{tabular}
	\end{adjustbox}
	\vspace{1mm}
\end{table*}

\begin{table*}[t]
	\caption{Download links and credits of the test set of the Sketchfab dataset \cite{qian2020pugeo}.}
	\label{table: sketchfab composition test}
	\vspace{-2mm}
	\centering
	\renewcommand{\arraystretch}{1.25}
	\begin{adjustbox}{max width=\linewidth}
		\begin{tabular}{llp{220mm}}
			\toprule
			\makecell{Name} &
			\makecell{Download link} &
			\makecell{Credit}  
			\\
			\midrule
			A9-vulcan\_aligned  & https://skfb.ly/6AAGM & "Vulcan" (https://skfb.ly/6AAGM) by Geoffrey Marchal is licensed under Creative Commons Attribution-NonCommercial (http://creativecommons.org/licenses/by-nc/4.0/). \\
			a72-seated\_jew\_aligned & https://skfb.ly/6ynCI & "Seated Jew" (https://skfb.ly/6ynCI) by Geoffrey Marchal is licensed under Creative Commons Attribution-NonCommercial (http://creativecommons.org/licenses/by-nc/4.0/). \\
			asklepios\_aligned & https://skfb.ly/6zt76 & "Young roman as Asklepios" (https://skfb.ly/6zt76) by Geoffrey Marchal is licensed under Creative Commons Attribution-NonCommercial (http://creativecommons.org/licenses/by-nc/4.0/). \\
			baron\_seutin\_aligned & https://skfb.ly/6Bq6Y & "Baron Seutin" (https://skfb.ly/6Bq6Y) by Geoffrey Marchal is licensed under Creative Commons Attribution-NonCommercial (http://creativecommons.org/licenses/by-nc/4.0/). \\
			charite\_-\_CleanUp\_-\_LowPoly\_aligned & https://skfb.ly/6yWCX & "The Charity" (https://skfb.ly/6yWCX) by Geoffrey Marchal is licensed under Creative Commons Attribution-NonCommercial (http://creativecommons.org/licenses/by-nc/4.0/). \\
			cheval\_terracotta\_-\_LowPoly-RealOne\_aligned & https://skfb.ly/6APDT & "Relief in terracotta" (https://skfb.ly/6APDT) by Geoffrey Marchal is licensed under Creative Commons Attribution-NonCommercial (http://creativecommons.org/licenses/by-nc/4.0/). \\
			cupid\_aligned & https://skfb.ly/6vN6Z & "Cupid" (https://skfb.ly/6vN6Z) by Geoffrey Marchal is licensed under Creative Commons Attribution-NonCommercial (http://creativecommons.org/licenses/by-nc/4.0/). \\
			dame\_assise\_-\_CleanUp\_-\_LowPoly\_aligned & https://skfb.ly/6BNFn & "Seated Lady" (https://skfb.ly/6BNFn) by Geoffrey Marchal is licensed under Creative Commons Attribution-NonCommercial (http://creativecommons.org/licenses/by-nc/4.0/). \\
			drunkard\_-\_CleanUp\_-\_LowPoly\_aligned & https://skfb.ly/6BH8D & "Drunkard - Cap Re" (https://skfb.ly/6BH8D) by Geoffrey Marchal is licensed under Creative Commons Attribution-NonCommercial (http://creativecommons.org/licenses/by-nc/4.0/). \\
			Gramme\_aligned & https://skfb.ly/6ABqV & "Statue of Zénobe Gramme" (https://skfb.ly/6ABqV) by LZ Creation is licensed under Creative Commons Attribution (http://creativecommons.org/licenses/by/4.0/). \\
			madeleine\_aligned & https://skfb.ly/6AtTy & "Marie-Madeleine" (https://skfb.ly/6AtTy) by Geoffrey Marchal is licensed under Creative Commons Attribution-NonCommercial (http://creativecommons.org/licenses/by-nc/4.0/). \\
			retheur\_-\_LowPoly\_aligned & https://skfb.ly/ooIRB & "Bust of a rhetorician\_Restored" (https://skfb.ly/ooIRB) by Digital\_Restoration is licensed under Creative Commons Attribution-NonCommercial (http://creativecommons.org/licenses/by-nc/4.0/). \\
			saint\_lambert\_aligned & https://skfb.ly/6ANWw & "Lambert de Maestricht and Liège" (https://skfb.ly/6ANWw) by Geoffrey Marchal is licensed under Creative Commons Attribution-NonCommercial (http://creativecommons.org/licenses/by-nc/4.0/). \\
			\bottomrule
		\end{tabular}
	\end{adjustbox}
	\vspace{1mm}
\end{table*}

\end{document}